\documentclass{article}

\usepackage[numbers]{natbib}
\usepackage[final,nonatbib]{nips_2018}

\usepackage[utf8]{inputenc} 
\usepackage[T1]{fontenc}    
\usepackage{hyperref}       
\usepackage{url}            
\usepackage{booktabs}       
\usepackage{amsfonts}       
\usepackage{amsmath}       
\usepackage{nicefrac}       
\usepackage{microtype}      

\usepackage[colorinlistoftodos,textsize=tiny]{todonotes}
\usepackage{float}
\usepackage{subfig}
\usepackage{soul}

\newcommand{\mapST}{F_{S \rightarrow T}}
\newcommand{\mapTS}{F_{T \rightarrow S}}
\newcommand{\lossGT}{\mathcal{L}_{GT}(f_T, \mapST, x_S, y_S)}
\DeclareMathOperator{\E}{\mathbb{E}}
\newcommand{\vect}[1]{\mathbf{#1}}

\title{GANtruth -- an unpaired image-to-image translation method for driving scenarios}

\author{
  Sebastian Bujwid \\ 
  KTH Royal Institute of Technology \\
  Univrses AB \\
  \texttt{bujwid@kth.se} \\
   \And
   Miquel Martí \\
   KTH Royal Institute of Technology \\
   Univrses AB \\
   \texttt{miquelmr@kth.se} \\
   \AND
   Hossein Azizpour \\
   KTH Royal Institute of Technology \\
   \texttt{azizpour@kth.se} \\
   \And
   Alessandro Pieropan \\
   Univrses AB \\
   \texttt{alessandro.pieropan@univrses.com} \\
}

\begin{document}

\maketitle

\begin{abstract}
Synthetic image translation has significant potentials in autonomous transportation systems. That is due to the expense of data collection and annotation as well as the unmanageable diversity of real-words situations. The main issue with unpaired image-to-image translation is the ill-posed nature of the problem. In this work, we propose a novel method for constraining the output space of unpaired image-to-image translation. We make the assumption that the environment of the source domain is known (e.g. synthetically generated), and we propose to explicitly enforce preservation of the ground-truth labels on the translated images.

We experiment on preserving ground-truth information such as semantic segmentation, disparity, and instance segmentation. We show significant evidence that our method achieves improved performance over the state-of-the-art model of UNIT for translating images from SYNTHIA to Cityscapes. The generated images are perceived as more realistic in human surveys and outperforms UNIT when used in a domain adaptation scenario for semantic segmentation.
\end{abstract}





\section{Introduction}
Autonomous driving has a high potential for improving human life in many aspects, such as reducing the number of accidents and giving commuters the choice to dedicate their travel time to other activities.
However, due to the diverse nature of the environment and the unpredictable behavior of humans we are still far from solving the problem and a particular effort is needed in order to enable a reliable understanding of the surroundings, essential for decision making. In this regard, deep learning techniques have shown impressive results in many tasks such as object detection, tracking and semantic segmentation. These methods, though, require vast amounts of data, which 
is very expensive to collect and annotate. An appealing solution lies in using synthetic data instead \cite{richter2016playing,Johnson17,mayer2016large}, given that the ground-truth annotation can be extracted automatically from the simulators.
Such an approach has shown promising results in applications like human pose estimation and hand tracking \cite{danielsson2014, real-time-human-pose-recognition-in-parts-from-a-single-depth-image}.

Synthetic data has enormous potential especially in autonomous driving, because safety critical situations that are exceptionally rare in real life can be simply simulated.
Such flexibility also allows to render data corresponding to various environments and conditions which make it easier to adapt cars to different countries where not only the environment, but also regulations might be different. Given the added value of synthetic data, our work aims to improve it's realism by finding a mapping between real and synthetic domains using generative models to translate unpaired sets of images.
However, since the problem is generally ill-posed and thus can have infinitely many solutions, in this work we aim to constrain the output space by the preservation of the content that should be representation invariant in the special case of synthetic-to-realistic translation and thereby improve the translation utility.
We enforce the preservation of synthetic domain ground-truth information in the translated images by means of a loss between the source domain ground-truth labels and the output of a pre-trained label estimator on the target domain.
The results of the method we propose are judged as more realistic by human evaluators and achieve better performance on a standard domain adaptation scenario than the state-of-the-art model of UNIT.

\section{Related work}
Image-to-image translation can be considered as a variation of the visual adaptation problem  firstly introduced by \citet{Saenko2010}. In recent years, using generative adversarial networks (GANs)~\cite{goodfellow2014gan} to synthesize images has shown promising results. The main idea lies in training two competing networks, while the generator learns to produce realistic images, the discriminator network learns to determine if an image is real or not.
However, in GANs the output depends on the input random noise vector leaving no semantic control on the generated data. A proposed solution to the former is to condition the output on the source domain samples 
leading to many interesting applications in image-to-image translation such as the generation of natural scenes~\cite{Wang_SSGAN2016}, super resolution of images~\cite{wang2017high} or image manipulation~\cite{zhu2016generative,Johnson2016Perceptual}. 
Some methods proposed in recent years build upon the assumption that images in the source domain have correspondences with the images in the target domain. With this assumption, \citet{Isola2017ImagetoImageTW} proposed a network architecture and loss function and produced impressive results. \citet{wang2017high} has shown that it is possible to generate high resolution images from semantic segmentation maps by learning the mapping between the domains.
Despite the impressive results of the mentioned methods, the assumption of having correspondences between the domains is quite limiting due to the scarcity of manually labeled data or the nature of the chosen domain pairs themselves, e.g. depth maps with natural images or synthetic images with natural images.
In order to overcome this limitation \citet{liu2017unsupervised} proposed forcing a shared latent space between the domains in the framework called UNIT. Moreover, they enforced cycle consistency as introduced in CycleGAN~\cite{zhu2017unpaired}, DiscoGAN~\cite{pmlr-v70-kim17a} and DualGAN~\cite{Yi_2017_ICCV_dualGAN}, so that a generated image in the target domain could be mapped back to the source domain.

In this work, we are interested in preserving the domain-agnostic content of the images.
Previous works in this direction relied on jointly learning the task network on the translated images \cite{Zheng_2018_ECCV,Hoffman_cycada2017}.
The problem with such an approach is that it does not necessarily have to force the translated images to be consistent with the source samples if only the task networks learn to ignore the content that was modified.
CyCADA \cite{Hoffman_cycada2017} is conceptually closest to our work in that it also tries to preserve semantic consistency during the translation. CyCADA proposes to minimize the discrepancy between labels estimated by a model trained on source domain. We argue that this approach can be problematic since it assumes the label estimators trained on the source domain is re-usable for the target domain. This assumption essentially violates the definition of the domain adaptation (\textit{i.e.} shift in the visual representation). 

\section{GANtruth}
\begin{figure}[t!]
	\centering
	\includegraphics[width=0.70\textwidth]{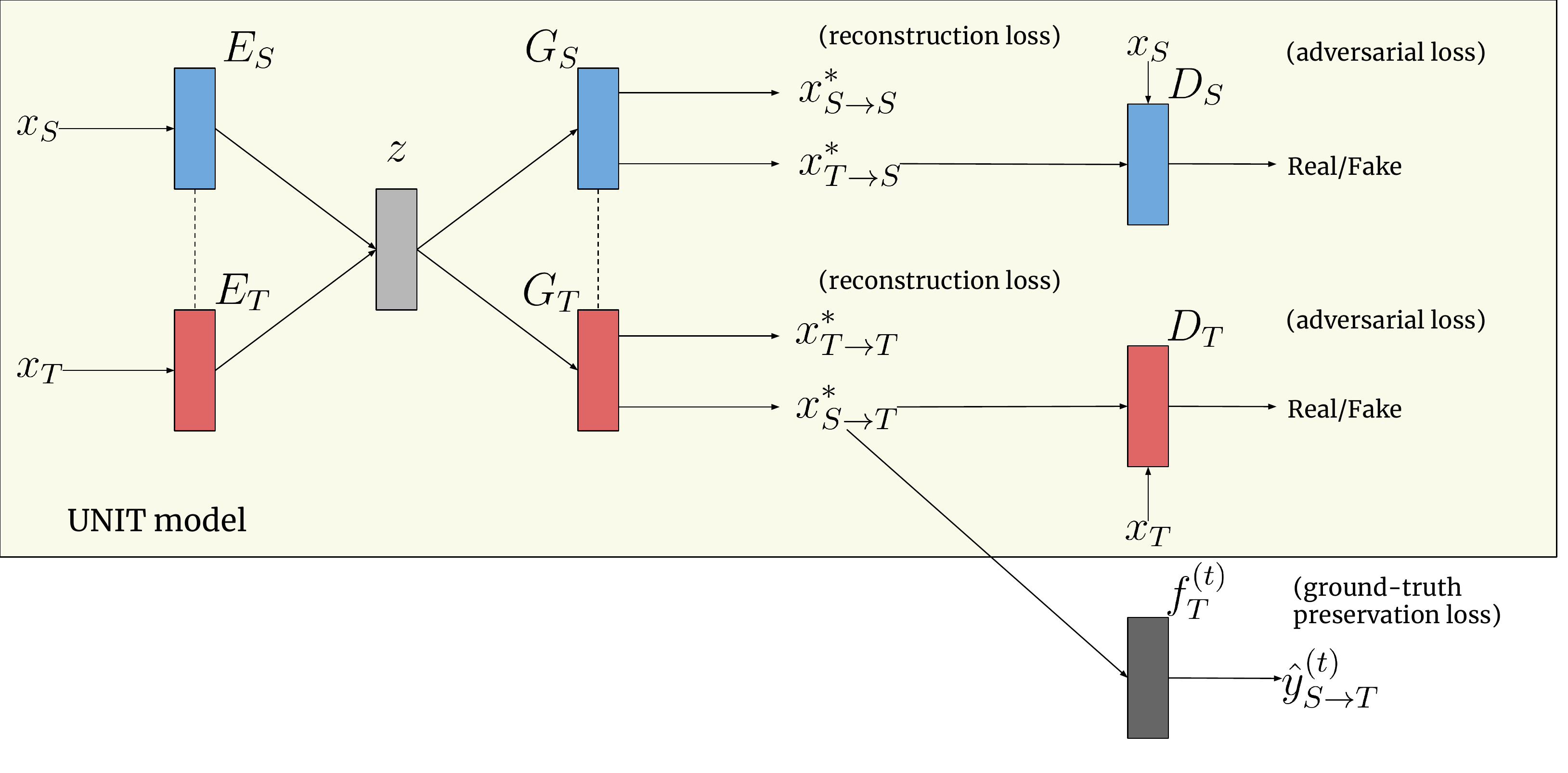}
	\caption{\textbf{Diagram of GANtruth based on UNIT framework (GANtruth+UNIT)}. The model takes as an input a sample from each domain and encodes the samples into the same latent space. Then, from the latent space, those samples can be translated back to the same domain using the decoder of the same domain as the original sample, or translated to the other domain using the other decoder. The blue modules work in the source domain, while the red ones in the target domain and dashed lines mean that weights are partially shared.}
	\label{fig:realgan_UNIT_diagram}
\end{figure}

Assume an image space $\mathbb{X}$ and two probability distributions $P_S(x)$ and $P_T(x)$ over $\mathbb{X}$ denoting the source and target image domains respectively. In unpaired image-to-image translation, given the training sets $X_S=\{x_S \sim P_S(x)\}$ and $X_T=\{x_T \sim P_T(x)\}$, we are interested in learning an adaptation function $\mapST$ such that $\mapST(x\sim P_S)\sim P_T$ \footnote{In its general form, the translation function can produce $P(x_T|x_S)$ instead of a deterministic mapping.} 

In general, this is an ill-posed problem where several valid solutions can exist. Also, in the particular case of synthetic-to-natural image translation many (potentially infinite) valid natural images might exist for a given synthetic image 
. Motivated by this problem, we propose to constrain the learning problem by requiring the image content to be preserved during the translation. Formally, assume a label space $\mathbb{Y}$, we require a translation function that additionally satisfies the following constraint.
\begin{equation}
P(y|x_S)=p(y|\mapST(x_S))\quad \forall x_S\sim P_S
\label{eq:label_preservation}
\end{equation}
This formulation assumes access to the label-conditional in both domains. We relax the constraint in the following \textit{ground-truth preservation loss} function:
\begin{equation} 
\mathcal{L}_{GT} = \sum_{x_S\in X_S} \mathcal{L}(f_T(\mapST(x_S)), y_S)
\label{eq:label_preservation_loss}
\end{equation}
where $\mathcal{L}$ is some distance measure over $\mathbb{Y}$. The loss function assumes labels for the source images ($y_S$). Our source images are synthetically generated which come with free information about the environment (used to render $x_S$). We also assume \textit{some} pre-trained estimator ($f_T$) for the target domain exists. In practice, some tasks are easier to solve than others, so we can leverage them for solving more difficult ones.  
Several ground-truth preservation losses can be aggregated if different pre-trained estimators ($f_{T}^{(t)}$) are available for the target domain. In fact, this is usually the case in synthetic-to-real images, since several off-the-shelf visual recognition (\textit{e.g.} object detection, semantic segmentation, depth estimation, instance segmentation) models are available for natural images.

\subsection{Translation Models}
We examine the effectiveness of the ground-truth preservation loss ($\mathcal{L}_{GT}$) on a basic GAN-based \cite{goodfellow2014gan} model, minimizing the following objective:
\begin{equation} \label{eq:realgan_loss_SA}
  \mathcal{L}_{GANtruth} = \lambda_{GAN} \cdot \mathcal{L}_{GAN} (E_S, G_T, D_T) + \lossGT.
\end{equation}
Additionally, we also study the impact of the ground-truth preservation loss on a state-of-the-art image-to-image translation model called UNIT \cite{liu2017unsupervised}.
UNIT is a bidirectional translation model which works through a shared latent space. The UNIT's translation functions consist of an encoder and decoder for each domain denoted by $E_T, E_S$ and $G_T, G_S$ respectively and is defined as $\mapST(x_S)=G_S (z \sim \mathcal{N} (E_S (x_S), I))$. UNIT's objective function augmented with the ground-truth preservation loss becomes:
\begin{equation}
\begin{aligned}
&\mathcal{L}_{GANtruth+UNIT} = \mathcal{L}_{VAE} (E_S, G_S) + \lambda_{GAN} \cdot \mathcal{L}_{GAN} (E_S, G_S, D_S) + \mathcal{L}_{CC} (E_S, G_S, E_T, G_T) \\
&+ \mathcal{L}_{VAE} (E_T, G_T) + \lambda_{GAN} \cdot \mathcal{L}_{GAN} (E_T, G_T, D_T) + \mathcal{L}_{CC} (E_T, G_T, E_S, G_S) \\
&+ \lossGT,
\end{aligned}
\end{equation}
with $D_T, D_S$ being discriminator functions that classify images into the two domains. The UNIT objective has several loss terms: Varitional AutoEncoder (VAE) \cite{Kingma2013AutoEncodingVB} losses denoted by $\mathcal{L}_{VAE}$ to ensure reconstruction from the latent space for source and target images, Generative Adversarial Network (GAN) \cite{goodfellow2014gan} losses denoted by $\mathcal{L}_{GAN}$ which try to make the translated source images indistinguishable from target images and vice versa, and cycle consistency \cite{zhu2017unpaired} losses (${L}_{CC}$) which additionally encourage perfect reconstruction of images after a translation to the other domain and back. In Figure \ref{fig:realgan_UNIT_diagram} we show a diagram of the augmented model. The parameters of $f_T$ are not updated during training. The details of the losses can be found in Appendix \ref{app:UNIT}.

\section{Experiments}
We experiment with our method using 16212 stereo images of \textit{Summer} and \textit{Spring} sequences from SYNTHIA-Sequences dataset~\cite{Ros_2016_CVPR} as a source domain.
As a target domain we use Cityscapes~\cite{cordts2016cityscapes} train and train-extra sets, containing 23472 images. The setup is similar to the one used by \citet{liu2017unsupervised}. The details of the network architecture are described in Appendix \ref{app:appendix_model_architecture}.

\subsection{Enabling preservation of different ground-truth}
We experiment with enforcing preservation of different types of labels.
Different modules corresponding to different ground-truth information are enabled in order to observe their individual impact, as well as the results of combining them together.

\paragraph{Semantic segmentation}
For preserving semantic information we use the semantic segmentation network ICNet \cite{zhao2017icnet} as $f_T$ and cross entropy loss between ground-truth source labels and the estimated labels on $\mapST(x_S)$.
In most of our experiments the model is pre-trained on Cityscapes training and validation set, except experiments in Section~\ref{sec:semseg_adaptation} where only training set was used to make the evaluation fair. Because of the differences between classes that are defined in our source and target datasets, we use the mapping defined in Appendix~\ref{app:label_mapping}.

\paragraph{Depth (disparity)}
For preserving depth information we use the monocular depth estimation network proposed by \citet{monodepth17}.
This is an unsupervised model that uses stereo frames during training and produces disparities from both images, which are used to reconstruct the corresponding left or right image.
The model is trained to minimize the error of image reconstruction, as well as to preserve consistency between disparities produced from the left and right image.
Finally, during test time, the model estimates disparities from just a single (mono) image.
The variant of the model we use is based on the ResNet-50 architecture and is trained on the Kitti dataset~\cite{Geiger2012CVPR}, which we believe is similar enough to our target domain dataset.
Predicted disparities are multiplied by a constant to compensate for different camera parameters. The discrepancy measure we use for depth preservation is mean absolute error.

\paragraph{Instance segmentation}
In order to preserve better fine details of specific objects relevant to autonomous driving scenarios, we enforce preservation of ground-truth labels in the instance segmentation task, which gives a different mask for each individual object in the scene. We choose a Mask-RCNN~\cite{He_2017_ICCV} instance segmentation network based on Inception-v2~\cite{szegedy2016rethinking} and use the losses defined in \cite{He_2017_ICCV}. Instead of using a model trained on Cityscapes, we use one trained on Microsoft-COCO~\cite{microsoft-coco-common-objects-in-context}, whose images should belong to a domain much closer to the target domain than the synthetic images of the source domain. See the mapping between classes across datasets in Appendix~\ref{app:label_mapping}.

\paragraph{Hyperparameters}
We use Adam optimizer with learning rate equal $0.0001$ and $\beta_1 = 0.5$, $\beta_2 = 0.999$ and batch size equal $1$. We weight the loss function using the following coefficients:
$\lambda_{sem. seg.} = 40.0$, $\lambda_{disp.} = 0.4$, $\lambda_{inst.seg.} = 1.0$, $\lambda_{GAN} = 10.0$.

\subsection{Human surveys on perceptual realism}
To measure perceptual quality of the translated images we use human judges.
We run human surveys in the form of A/B tests, using the Amazon Mechanical Turk platform.
The participants are presented with two images and are asked to choose which one looks more realistic.
The images we show are outputs of different models from synthetic to real translation.
This way we measure a relative improvement of image quality against a baseline model.

The input images chosen for the survey are sampled randomly from the dataset and are the same for all the studies.
The order of the images (left vs. right) is selected randomly for each comparison.

\begin{table}[tb!]
    \begin{minipage}{.60\textwidth}
    \caption{AMT survey results, 30 images randomly sampled from the dataset are compared by 15 participants (450 responses) each. Numbers are average values. S = Semantic Segmentation, D = Depth, I = Instance Segmentation, S-SEQ = SYNTHIA-Seq, S-RAND = SYNTHIA-RAND-CVPR16}
    \label{tbl:amt_results}
    \centering
        \begin{tabular}{lcc}
  \toprule
   \multicolumn{3}{c}{Preference against UNIT \cite{liu2017unsupervised}} \\
  \cmidrule(r){2-3}
  Method & S-SEQ & S-RAND \\
  \midrule
  Simple GAN (baseline)             & 25.11\%          &   12.22\%    \\
  \midrule
  GANtruth (S)                      & 61.33\%          &   23.56\%    \\
  GANtruth (D)                      & 30.22\%          &    8.44\%    \\
  GANtruth (I)                      & 38.44\%          &   23.56\%    \\
  GANtruth (S+D)                    & \textbf{61.78\%} &   \textbf{59.11\%}    \\
  \midrule
  UNIT + GANtruth (S)               & 64.00\%          &  \textbf{78.89\%} \\
  UNIT + GANtruth (D)               & 57.55\%          &  43.33\%     \\
  UNIT + GANtruth (S+D)             & \textbf{70.22\%} &  70.00\%     \\
  UNIT + GANtruth (S+D+I)           & 65.78\%          &  61.11\%     \\
  \bottomrule
        \end{tabular}
    \end{minipage}
    \hspace{0.5cm}
    \begin{minipage}{.35\textwidth}
    \caption{Adaptation for semantic segmentation. Models are evaluated on Cityscapes validation set. S = Semantic Segmentation, D = Depth}
    \label{tbl:semseg_evaluation}
    \centering
        \begin{tabular}{lc}
  \toprule
  Dataset                                  & mIOU (\%)       \\
  \midrule
  Source domain \\ (SYNTHIA-Seq)              & 24.9            \\
  \midrule
  Simple GAN \\(baseline)                    & 20.3            \\
  UNIT \\(baseline)                          & 23.0            \\
  \midrule
  GANtruth (S+D)                           & \textbf{26.6}   \\
  \midrule
  Target domain \\ (Cityscapes-train set)     & 57.0            \\
  \bottomrule
        \end{tabular}
    \end{minipage} 
\end{table}

The results in Table~\ref{tbl:amt_results} show that images translated by GANtruth with both semantic segmentation and depth preservation are perceived as more realistic than UNIT on average, for images from both datasets. In the case of using only one of the modules available results are only comparable to UNIT in the case of semantic segmentation, but only on the SYNTHIA-Seq dataset. This can indicate that the different modules working at the same time help improving the generalization of the model. Furthermore, GANtruth improves humans' preference on all but one case when combining it with UNIT over only using UNIT, showing that the approaches are complementary.

\subsection{Qualitative results}
In Figure \ref{fig:qualitative_results} we show qualitative results for the different single-task GANtruth experiments, the combination of all considered tasks, and the later together with UNIT. In addition, we provide results for two baselines - a simple GAN and UNIT - for comparison.
\def\ganwidth{0.15\textwidth}
\begin{figure}[tb!]
  \centering
 \centering
 \begin{tabular}{ccc}
   & SYNTHIA-Seq & SYNTHIA-RAND-CVPR16 \\
  \rotatebox{90}{Source image} &
  \includegraphics[width=\ganwidth]{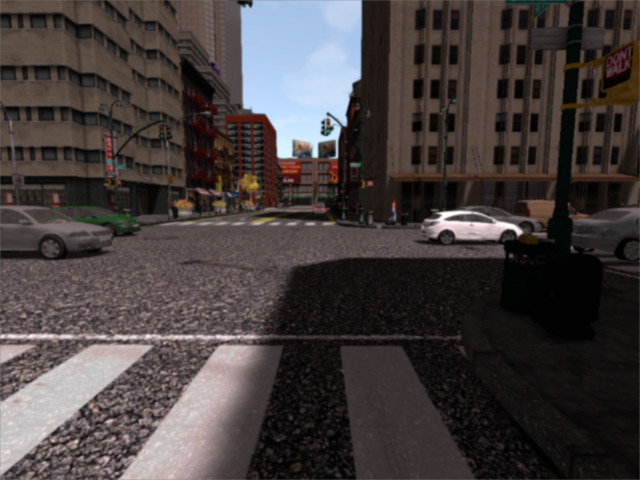}
  \includegraphics[width=\ganwidth]{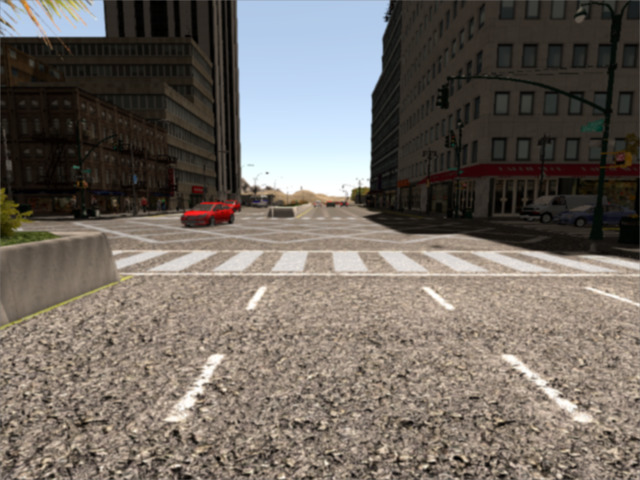} &
  \includegraphics[width=\ganwidth]{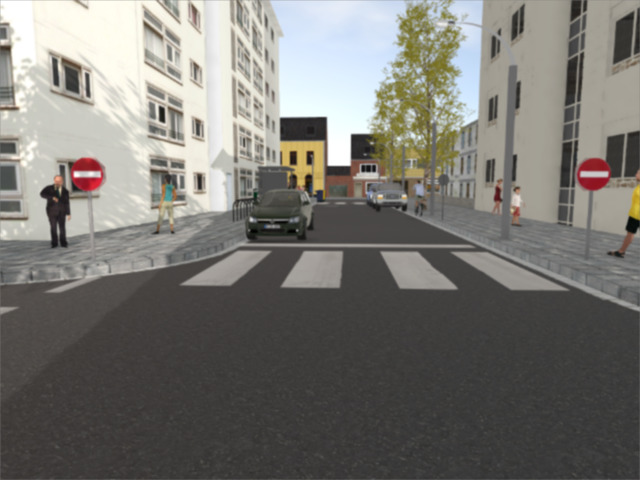} 
  \includegraphics[width=\ganwidth]{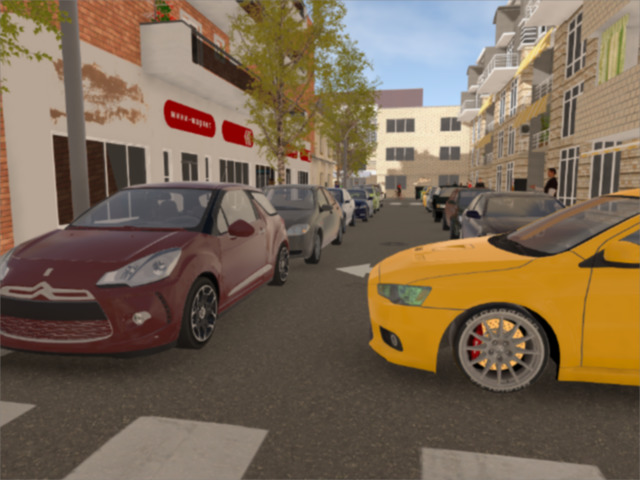} \\
  \rotatebox{90}{Simple} \rotatebox{90}{GAN} \rotatebox{90}{(baseline)} &
  \includegraphics[width=\ganwidth]{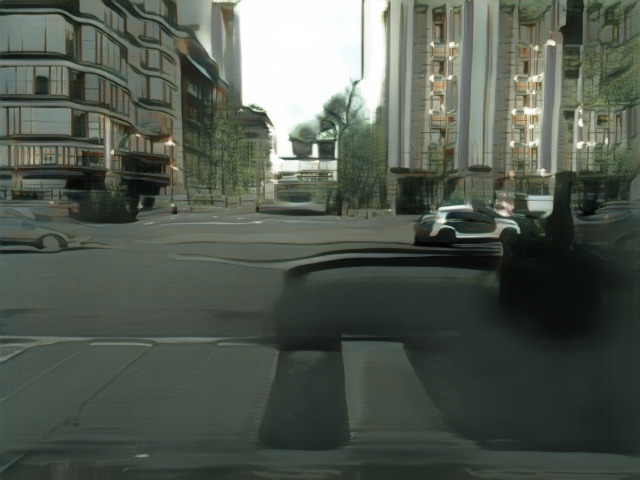}
  \includegraphics[width=\ganwidth]{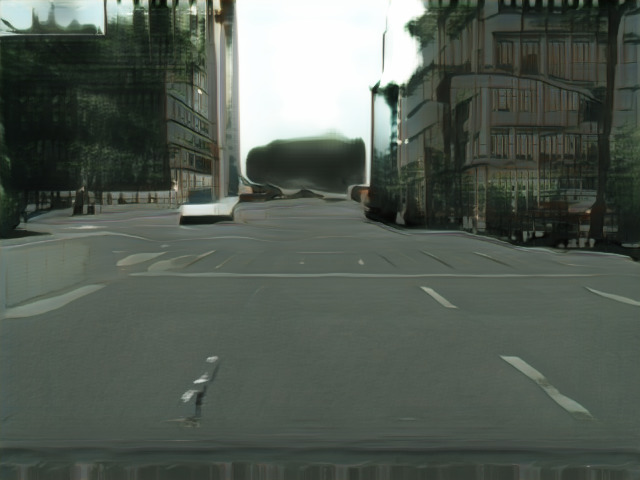} &
  \includegraphics[width=\ganwidth]{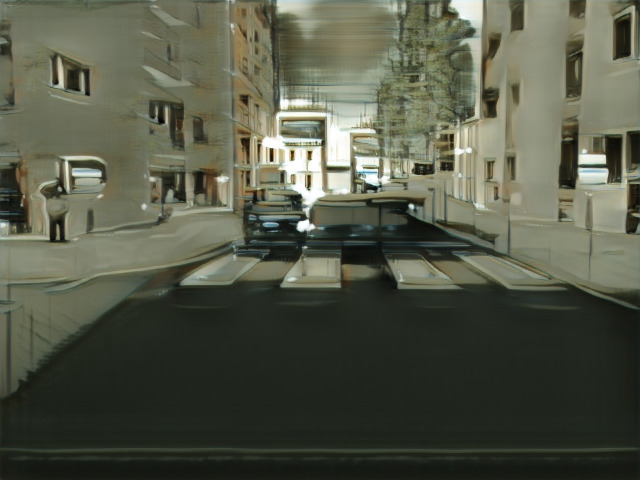} 
  \includegraphics[width=\ganwidth]{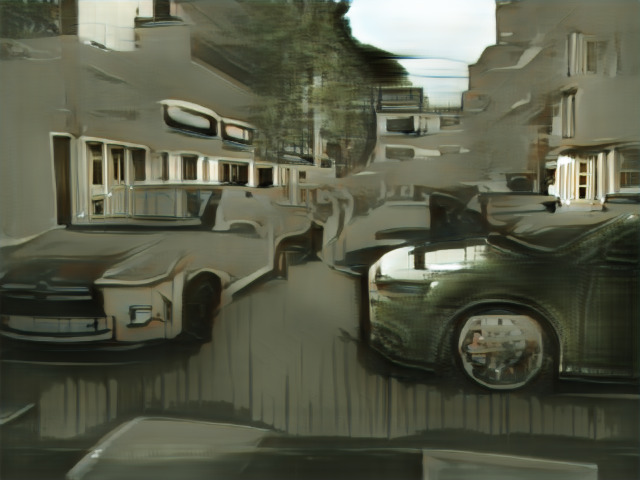} \\ 
  \rotatebox{90}{GANtruth} \rotatebox{90}{(S)} &   \includegraphics[width=\ganwidth]{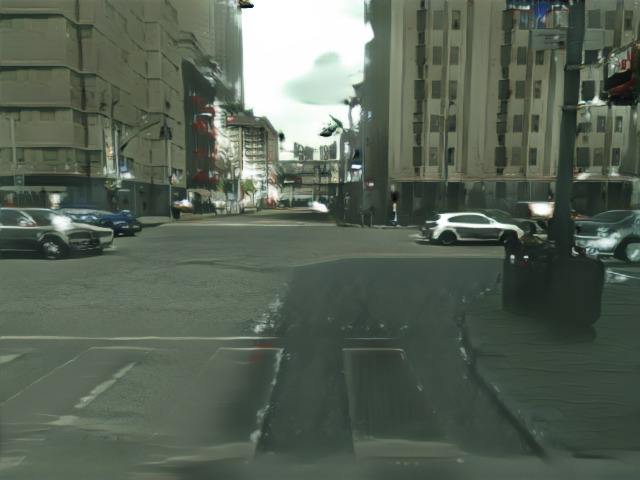}
  \includegraphics[width=\ganwidth]{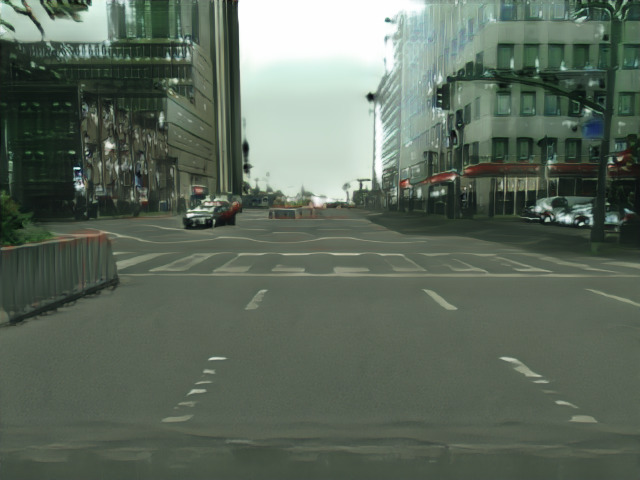} &
  \includegraphics[width=\ganwidth]{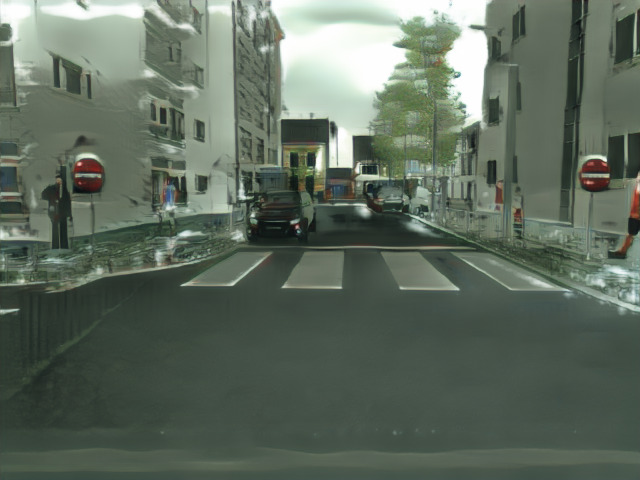} 
  \includegraphics[width=\ganwidth]{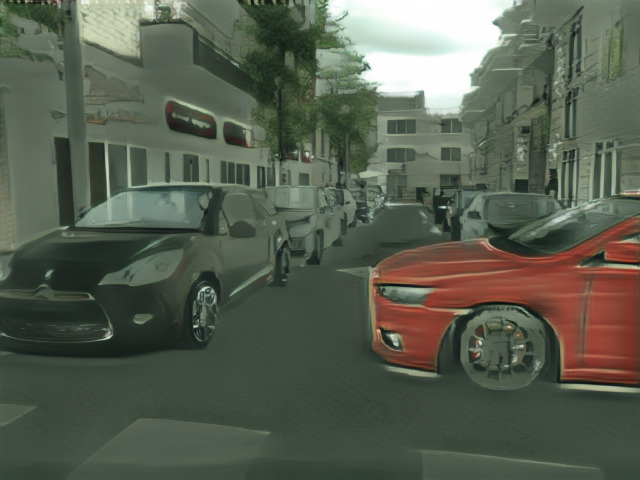} \\
  \rotatebox{90}{GANtruth} \rotatebox{90}{(D)} & \includegraphics[width=\ganwidth]{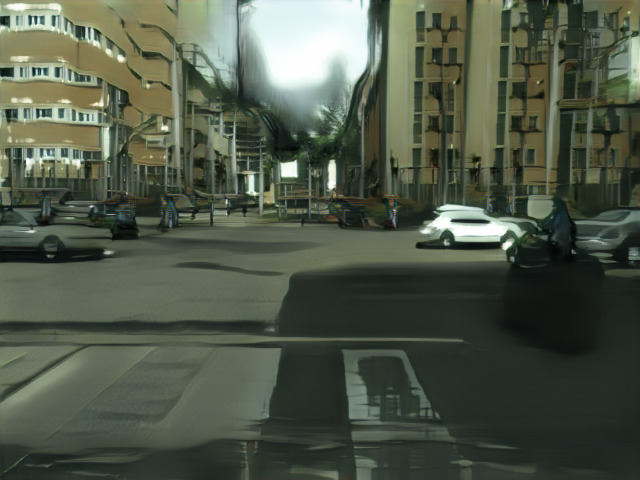}
  \includegraphics[width=\ganwidth]{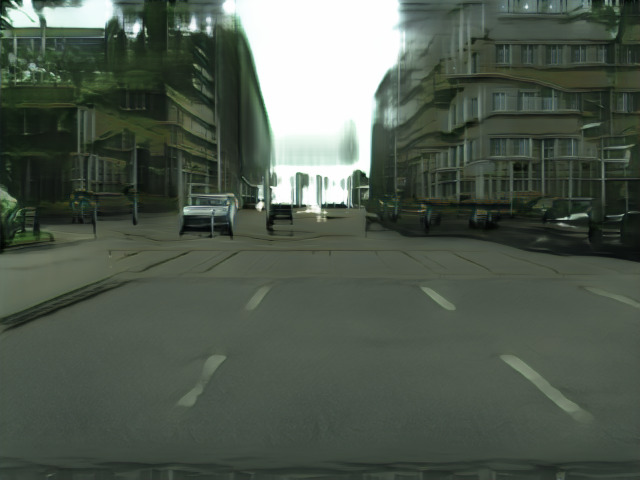} &
  \includegraphics[width=\ganwidth]{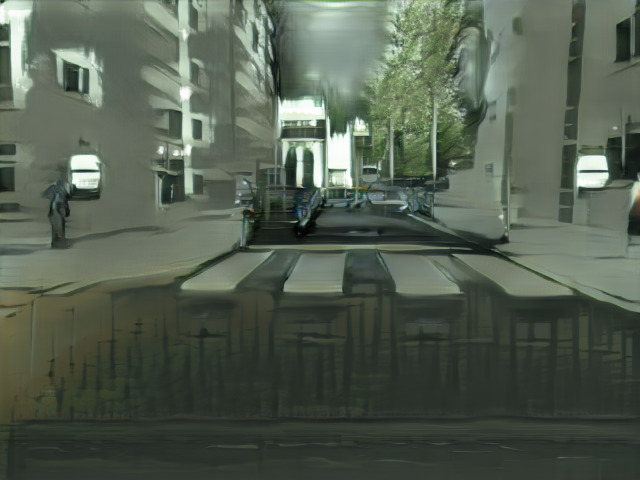} 
  \includegraphics[width=\ganwidth]{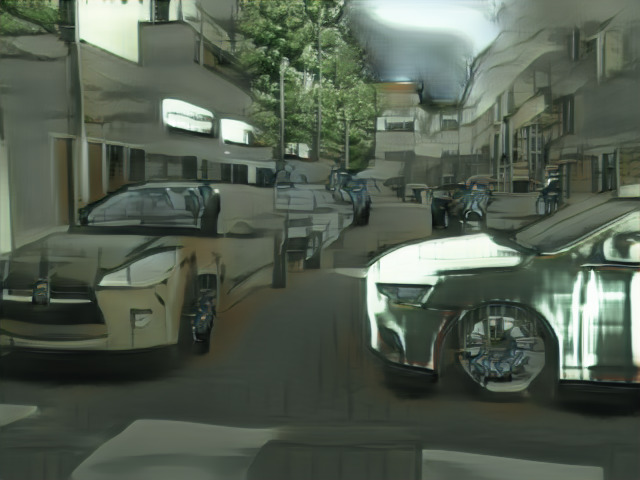} \\
  \rotatebox{90}{GANtruth} \rotatebox{90}{(I)} & \includegraphics[width=\ganwidth]{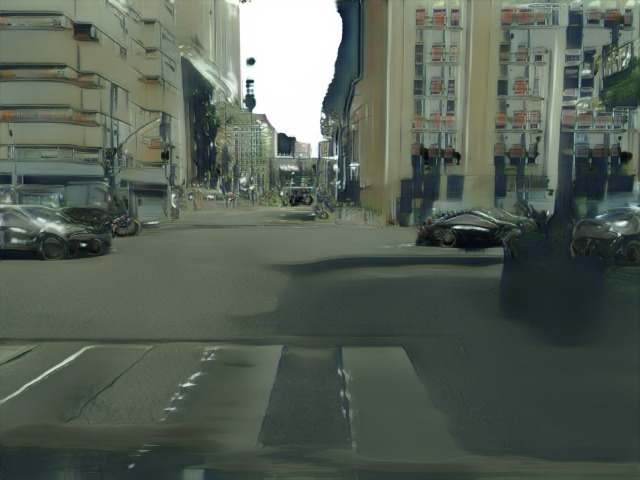}
  \includegraphics[width=\ganwidth]{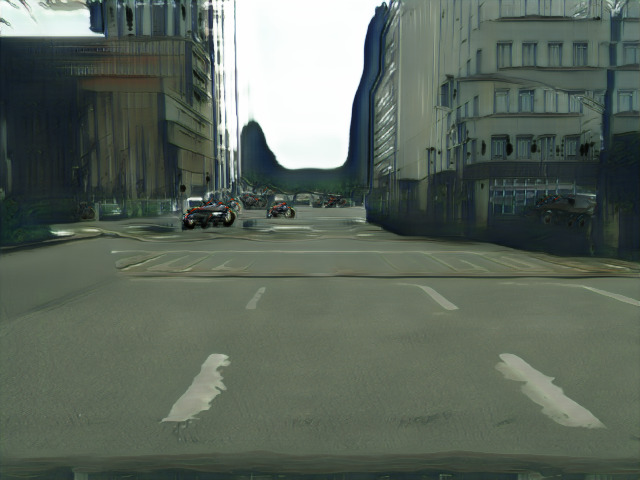} &
  \includegraphics[width=\ganwidth]{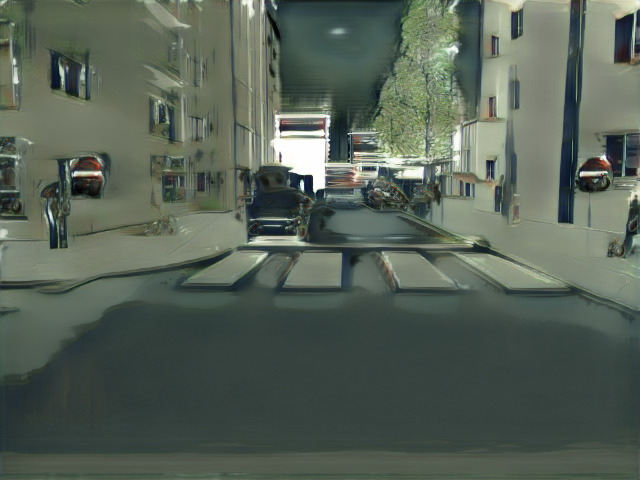} 
  \includegraphics[width=\ganwidth]{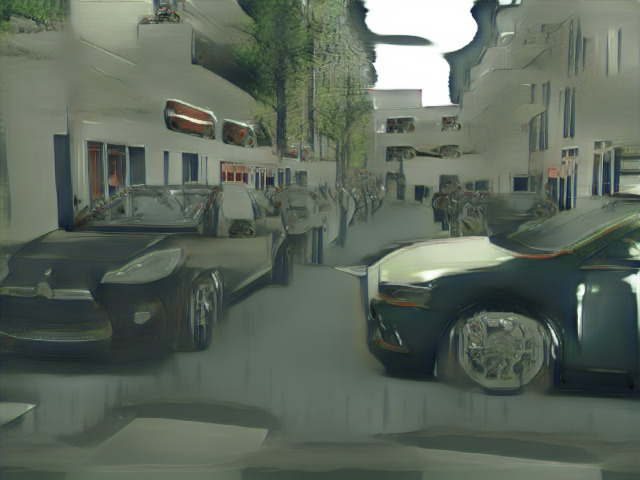} \\
  \rotatebox{90}{GANtruth} \rotatebox{90}{(S+D+I)} &
  \includegraphics[width=\ganwidth]{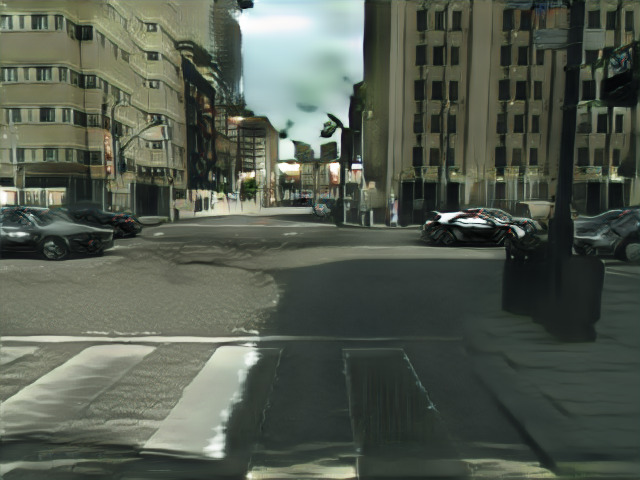}
  \includegraphics[width=\ganwidth]{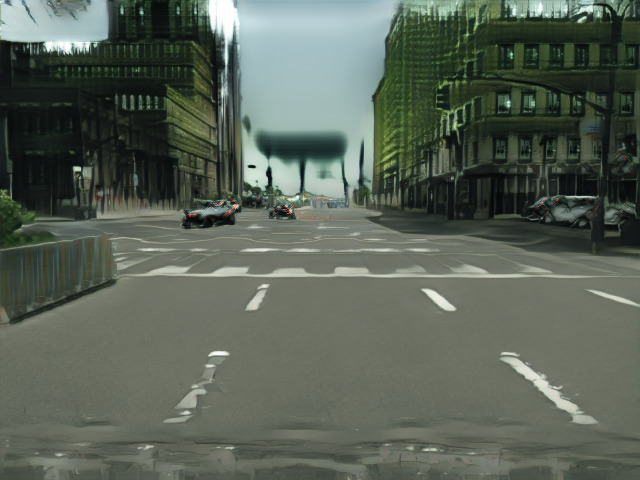} &
  \includegraphics[width=\ganwidth]{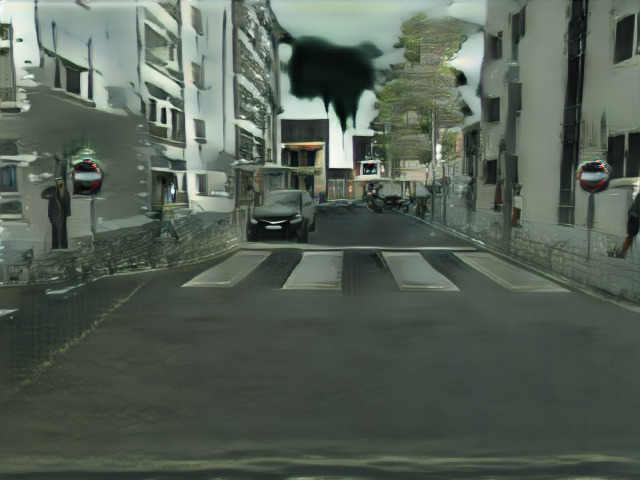} 
  \includegraphics[width=\ganwidth]{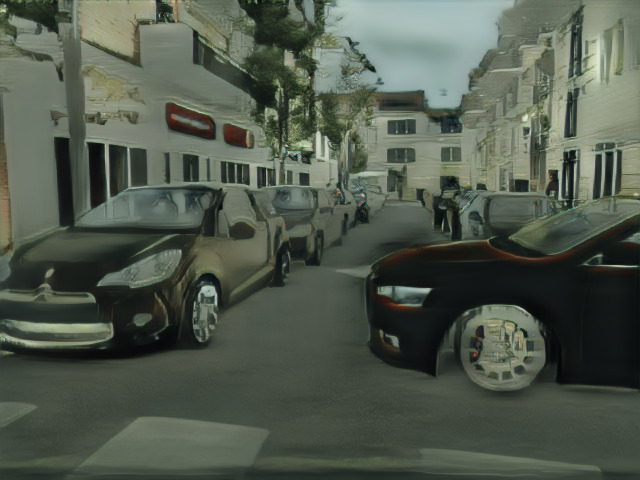} \\
  \rotatebox{90}{UNIT} \rotatebox{90}{(baseline)} &
  \includegraphics[width=\ganwidth]{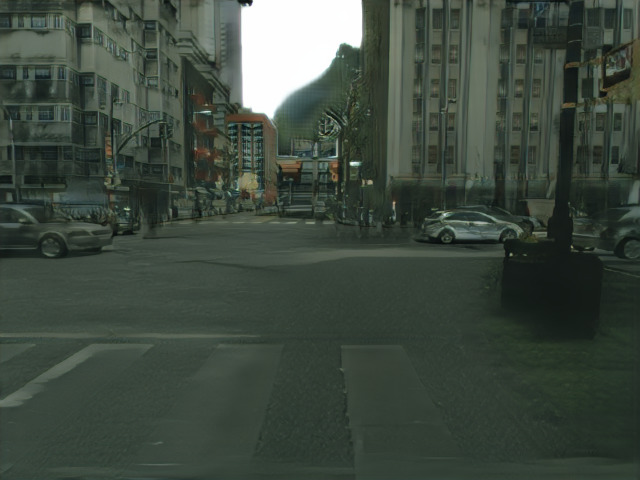}
  \includegraphics[width=\ganwidth]{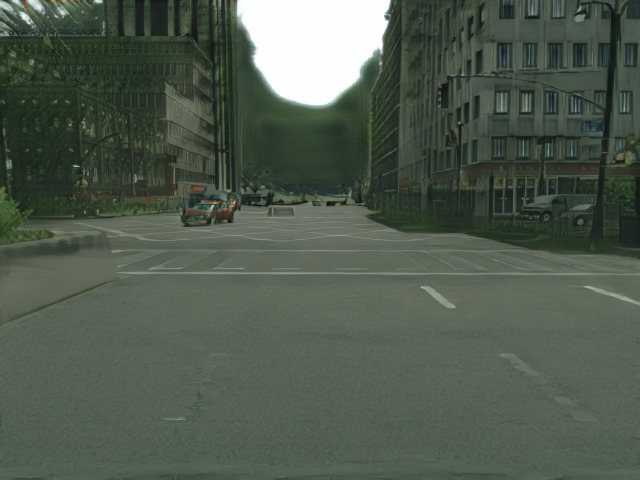} &
  \includegraphics[width=\ganwidth]{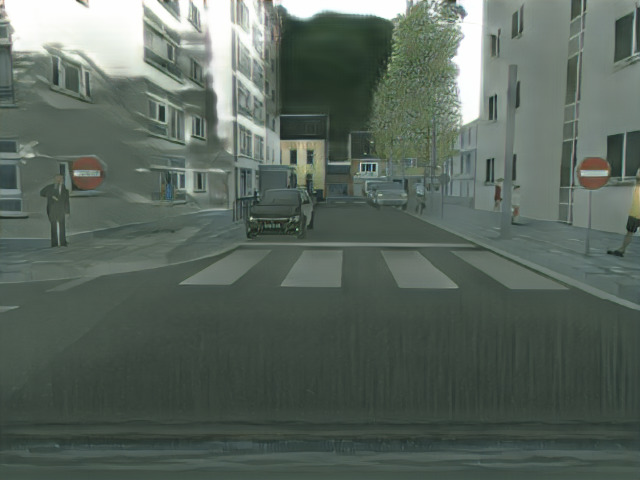} 
  \includegraphics[width=\ganwidth]{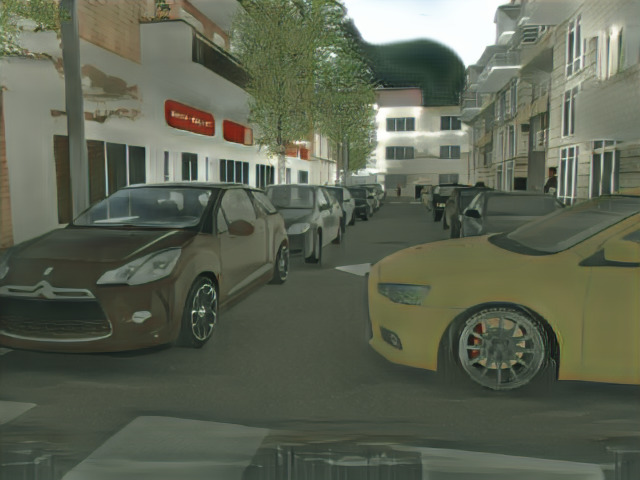} \\
  \rotatebox{90}{UNIT+} \rotatebox{90}{GANtruth} \rotatebox{90}{(S+D+I)}&
  \includegraphics[width=\ganwidth]{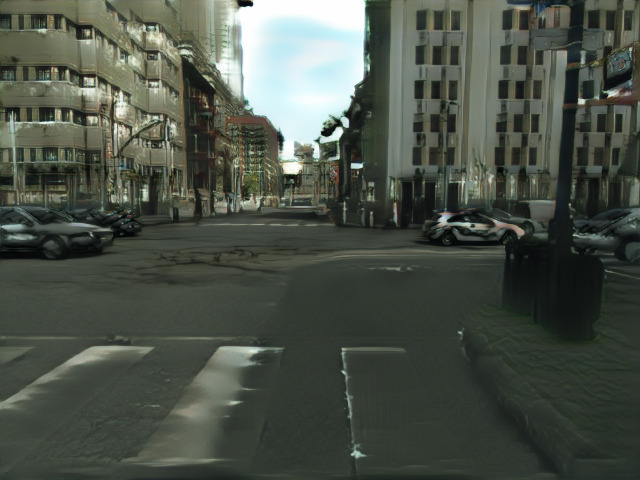}
  \includegraphics[width=\ganwidth]{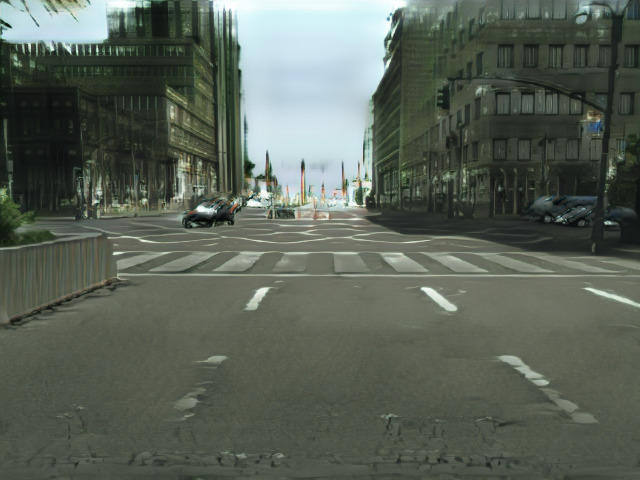} &
  \includegraphics[width=\ganwidth]{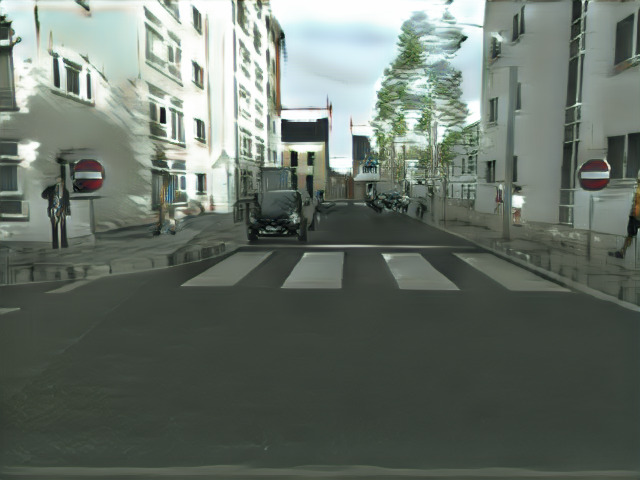} 
  \includegraphics[width=\ganwidth]{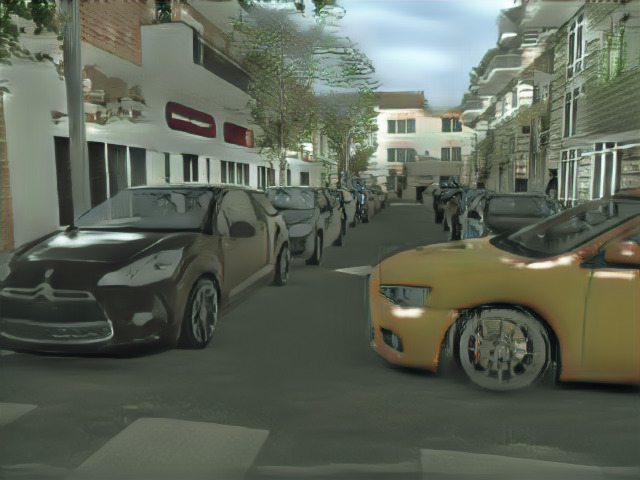} \\
 \end{tabular}
 \caption{Qualitative results. Images randomly sampled. S = Semantic Segmentation, D = Depth, I = Instance Segmentation.}
 \label{fig:qualitative_results}
\end{figure}

The superior performance of GANtruth with semantic segmentation labels preservation is confirmed in the qualitative results. Results of this method appear free of the largest artifacts in the baseline methods: objects melting into others in simple GAN, black sky in UNIT and simple GAN, vegetation on buildings in UNIT. However, colours of the source domain are not kept. The combination of GANtruth with UNIT keeps source image colours while removing the largest artifacts in UNIT.
More images are shown in Appendix \ref{app:additional_images}.

\subsection{Adaptation for semantic segmentation}
\label{sec:semseg_adaptation}

We evaluate our approach in a simple, two-step adaptation scenario. First, we construct new datasets made of synthetic data translated using different $\mapST$ models and corresponding ground-truth labels $y$.
Then, we use these datasets to train a supervised task network for the semantic segmentation problem.
Unlike other approaches \cite{Hoffman_cycada2017,Zheng_2018_ECCV}, our adaptation scenario uses two-separate steps -- we do not learn the task network during adaptation.

The task model we use to train on adapted images is DeepLabv3+ \cite{deeplabv3plus2018} with Xception \cite{chollet2017xception} backbone and output stride of 32. We fine-tune the model previously pre-trained on ImageNet for 90000 updates on 449x449 image crops with batch size 16. Batch normalization parameters are not updated. We find that these values give good results on the target domain dataset and we use it for all experiments.
In Table~\ref{tbl:semseg_evaluation} we present mIOU (mean Intersection-over-Union) scores on unseen Cityscapes validation set and show that training on translated images improves performance over training on synthetic data.

\subsection{Additional observations}

During our experimentation we noticed that applying image-to-image translation methods does not guarantee improvement on a simple, two-step domain adaptation.
This is the case for both baseline models, as well as some variants of our model.
Even though it is likely that the results would differ if a more advanced adaptation method was used (e.g. based on aligning feature distributions of the task network) instead of simply using the translated images, we believe that this is an interesting observation.
It suggests that the translation models despite producing visually appealing images might not necessarily be directly useful for training machine learning models as it was shown by \citet{shrivastava2017learning} on datasets that have much smaller domain gaps.
Additionally, we observe that when our semantic preservation loss is used, adding semantic consistency loss as proposed in CyCADA \cite{Hoffman_cycada2017} degrades the results.
This behaviour is somehow expected, since their approach to enforce semantic consistency uses a source domain estimator which cannot be expected to work well on images resembling the target domain.
For more details, please refer to Appendix \ref{app:additional_adaptation_results}.

\section{Conclusions}
In this work, we propose a novel approach for unpaired synthetic-to-realistic translation method on driving scenarios.
Our method based on explicitly enforcing preservation of ground-truth information is shown to be an effective approach for constraining the output space of the ill-posed problem of unpaired image-to-image translation.
Quantitative results from human surveys indicate that our model using both semantic segmentation and depth information preservation is perceived as more realistic than UNIT, but also that the two approaches are complementary and when combined produce even better results.
Additionally, we show that our method can be used for domain adaptation and using translated images for training semantic segmentation network improves the performance over using the original synthetic images.



\bibliographystyle{plainnat}
\bibliography{ref}
\newpage

\appendix

\section{UNIT}
\label{app:UNIT}
Instead of explicitly defining the relationship between images in different domains, UNsupervised Image-to-image Translation (UNIT) \cite{liu2017unsupervised} proposed to create a shared, domain-agnostic latent space.
Based on a previously introduced Coupled GAN (CoGAN) \cite{liu2016coupled}, they proposed an architecture consisting of two Variational Auto-encoders (VAEs) \cite{Kingma2013AutoEncodingVB} and two GANs.
The model simultaneously learns to translate in two directions.
Each encoder part of the VAEs corresponds to a different domain and both of them encode images into the same, shared latent space ($z$--space).
Additionally, decoder parts of each VAEs are appended with separate discriminators for given domains.
Combining those parts gives a separate GAN for each domain.

The framework allows to transform images in different directions, e.g. an image in domain $A$ is encoded into a $z$--space and after that can be decoded either into domain $B$ or back again into domain $A$.
The whole model is jointly trained to reconstruct the images with VAEs, translate images with GANs and additionally preserve cycle-consistency.

The UNIT's translation functions consist of an encoder and decoder for each domain denoted by $E_T, E_S$ and $G_T, G_S$ respectively and is defined as,

\begin{equation}
\begin{aligned}
& x^*_{S \rightarrow T} = \mapST (x_S) = G_T (z \sim \mathcal{N} (E_S (x_S), I)) \\
& x^*_{T \rightarrow S} = \mapTS (x_T) = G_S (z \sim \mathcal{N} (E_T (x_T), I)).
\end{aligned}
\end{equation} 

The objective function of UNIT consists of various terms and is given by:

\begin{equation}
\begin{aligned}
\mathcal{L}_{UNIT} = & \mathcal{L}_{VAE} (E_S, G_S) + \lambda_{GAN} \cdot \mathcal{L}_{GAN} (E_S, G_S, D_S) \\
& + \mathcal{L}_{CC} (E_S, G_S, E_T, G_T) \\
& + \mathcal{L}_{VAE} (E_T, G_T) + \lambda_{GAN} \cdot \mathcal{L}_{GAN} (E_T, G_T, D_T) \\
& + \mathcal{L}_{CC} (E_T, G_T, E_S, G_S)
\end{aligned}
\end{equation}

with $D_T, D_S$ being discriminator functions that classify images into the two domains. Generative Adversarial Network (GAN) \cite{goodfellow2014gan} losses denoted by $\mathcal{L}_{GAN}$ try to make the translated source images indistinguishable from target images and vice versa. The GAN losses are defined as:

\begin{equation} \label{eq:gan_non_saturating_loss}
\begin{aligned}
& \mathcal{L}_{GAN} = \mathcal{L}_{GAN}^{(D)} + \mathcal{L}_{GAN}^{(G)} \\
& \mathcal{L}_{GAN}^{(D)} = -\frac{1}{2} \E_{\vect{x} \sim p_{data}} \log D(\vect{x}) - \frac{1}{2} \E_{\vect{z}} \log (1-D(G(\vect{z}))) \\
& \mathcal{L}_{GAN}^{(G)} = -\frac{1}{2} \E_{\vect{z}} \log(D(G(\vect{z}))).
\end{aligned}
\end{equation}

Varitional AutoEncoder (VAE) \cite{Kingma2013AutoEncodingVB} losses denoted by $\mathcal{L}_{VAE}$ ensure reconstruction from the latent space for source and target images and is formally defined by

\begin{equation}
\begin{aligned}
\mathcal{L}_{VAE} (E, G) = &
\lambda_{kl} \cdot D_{KL} (\mathcal{N} (E(x), I) \|\; \mathcal{N} (0, I)) \\
& - \lambda_{ll} \E_{z \sim \mathcal{N} (E(x), I)} [\log p_G (x | z)]
\end{aligned}
\end{equation}

Cycle consistency \cite{zhu2017unpaired} losses (${L}_{CC}$) additionally encourage perfect reconstruction of images after a translation to the other domain and back. The cycle-consistency loss is defined as:

\begin{equation}
\begin{aligned}
\mathcal{L}_{CC} & (E_S, G_S, E_T, G_T) = \lambda_{kl} \cdot D_{KL} (\mathcal{N}( E_S(x_S), I) \|\; \mathcal{N} (0, I)) \\
& + \lambda_{kl} (\mathcal{N} (E_T( G_T (z \sim \mathcal{N}(E_S(x_s), I)), I ) \|\; \mathcal{N} (0, I)) \\
& - \lambda_{ll} \E_{z \sim \mathcal{N} ( E_T (G_T (z \sim \mathcal{N} (E_S (x_S), I) ) , I)} [\log p_{G_{T}} (x_S | z)]
\end{aligned}
\end{equation}

Finally, it is worth mentioning, that the weights of the last layers of the encoders and the first layers of the decoders are shared between domains. This helps to preserve a universal semantic meaning of the latent space.

\section{Additional images}
\label{app:additional_images}

\def\ganwidthapp{0.205\textwidth}
\begin{figure}[H]
  \centering
 \centering
 \begin{tabular}{ccc}
   & SYNTHIA-Seq & SYNTHIA-RAND-CVPR16 \\
  \rotatebox{90}{Source image} &
  \includegraphics[width=\ganwidthapp]{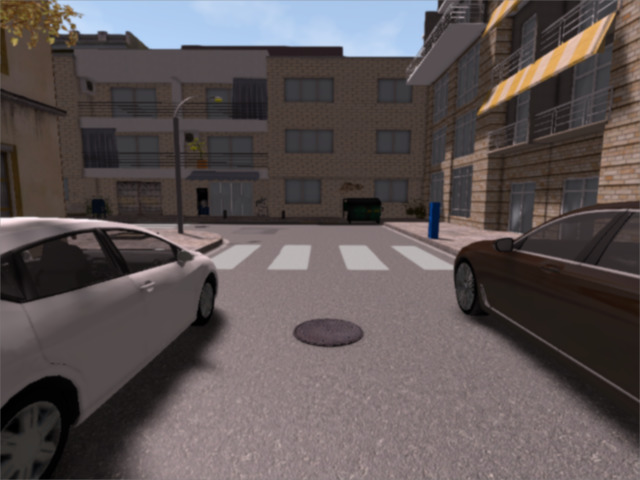}
  \includegraphics[width=\ganwidthapp]{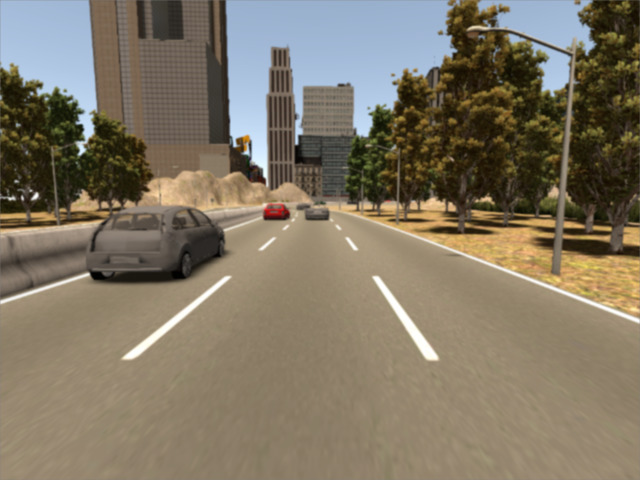} &
  \includegraphics[width=\ganwidthapp]{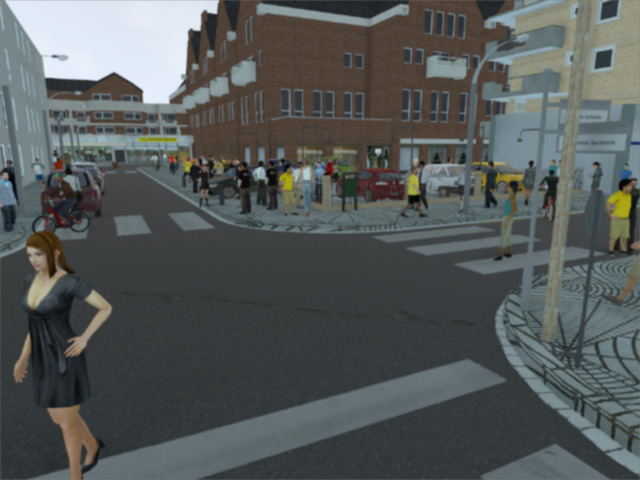} 
  \includegraphics[width=\ganwidthapp]{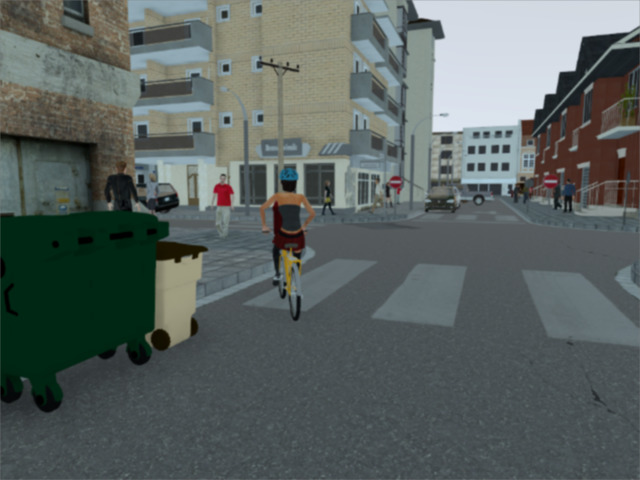} \\
  \rotatebox{90}{Simple} \rotatebox{90}{GAN} \rotatebox{90}{(baseline)} &
  \includegraphics[width=\ganwidthapp]{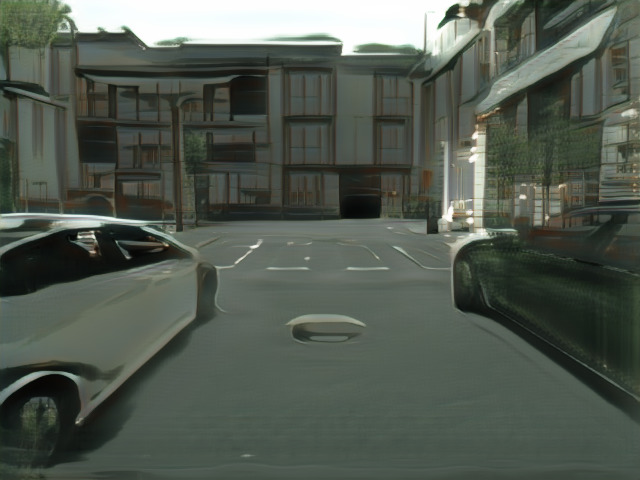}
  \includegraphics[width=\ganwidthapp]{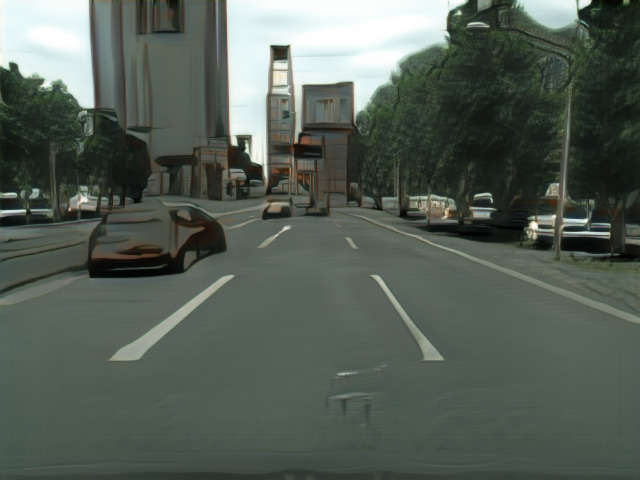} &
  \includegraphics[width=\ganwidthapp]{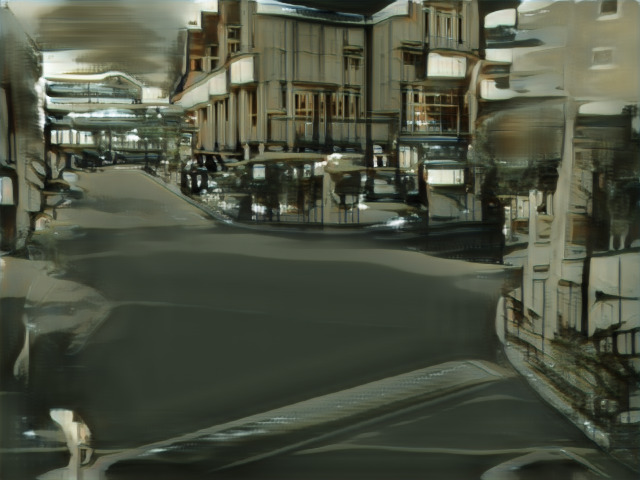} 
  \includegraphics[width=\ganwidthapp]{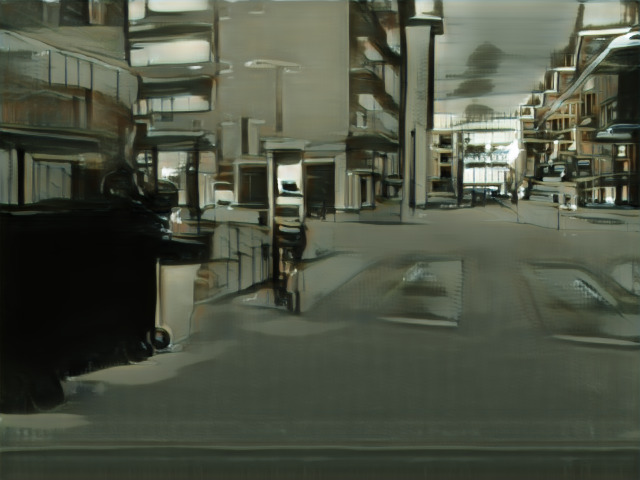} \\ 
  \rotatebox{90}{GANtruth} \rotatebox{90}{(S)} &   \includegraphics[width=\ganwidthapp]{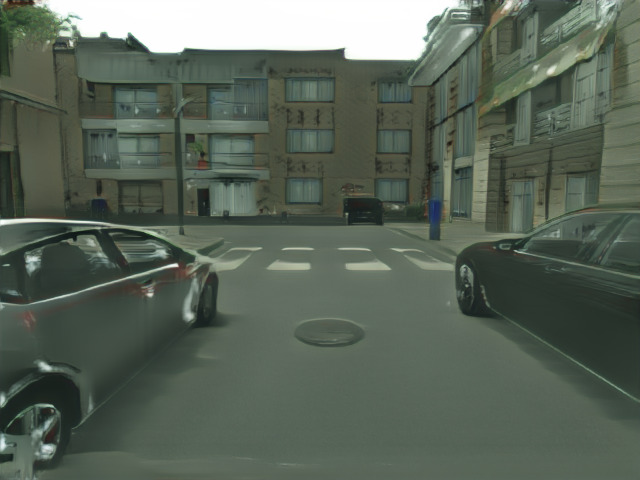}
  \includegraphics[width=\ganwidthapp]{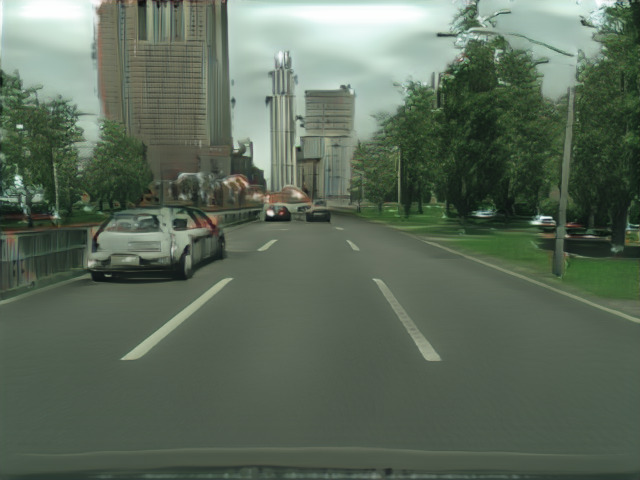} &
  \includegraphics[width=\ganwidthapp]{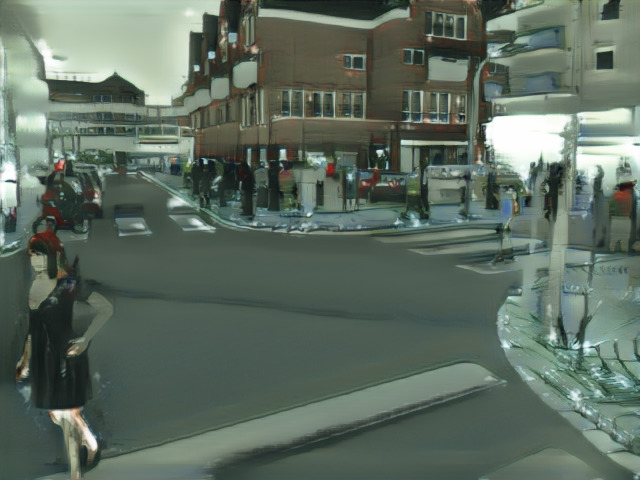} 
  \includegraphics[width=\ganwidthapp]{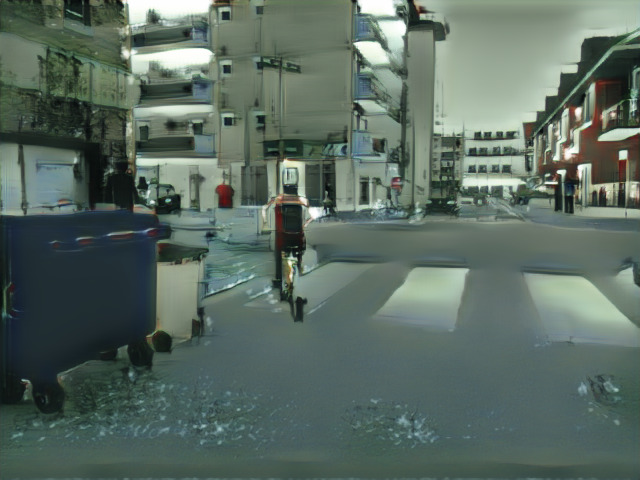} \\
  \rotatebox{90}{GANtruth} \rotatebox{90}{(D)} & \includegraphics[width=\ganwidthapp]{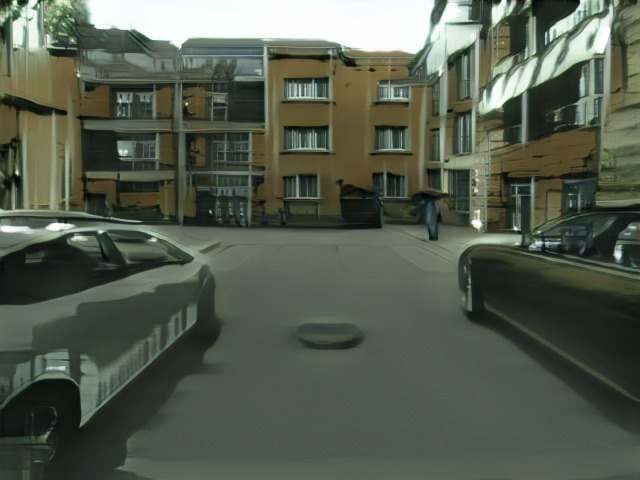}
  \includegraphics[width=\ganwidthapp]{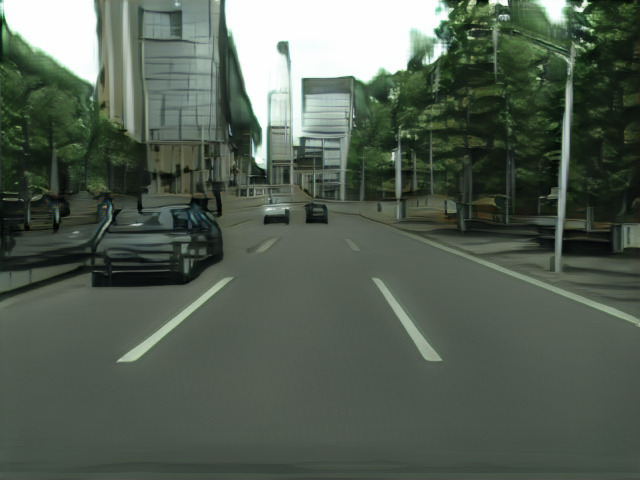} &
  \includegraphics[width=\ganwidthapp]{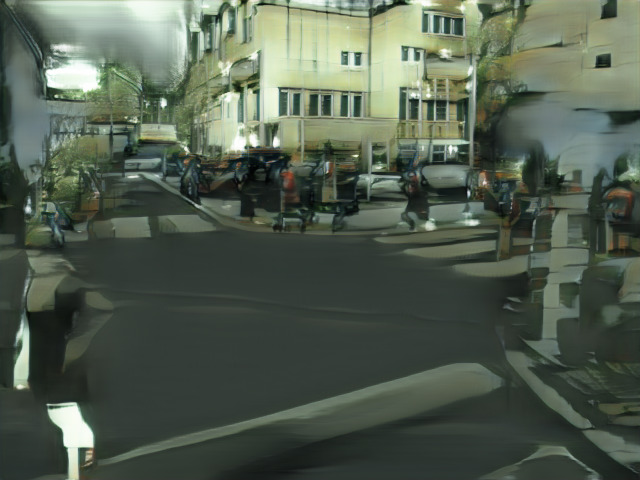} 
  \includegraphics[width=\ganwidthapp]{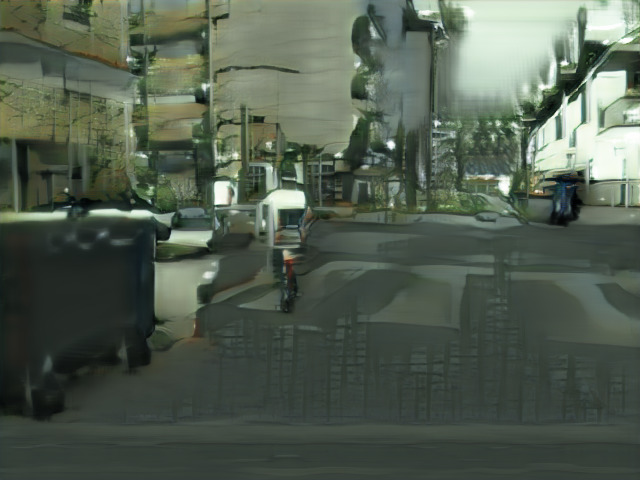} \\
  \rotatebox{90}{GANtruth} \rotatebox{90}{(I)} & \includegraphics[width=\ganwidthapp]{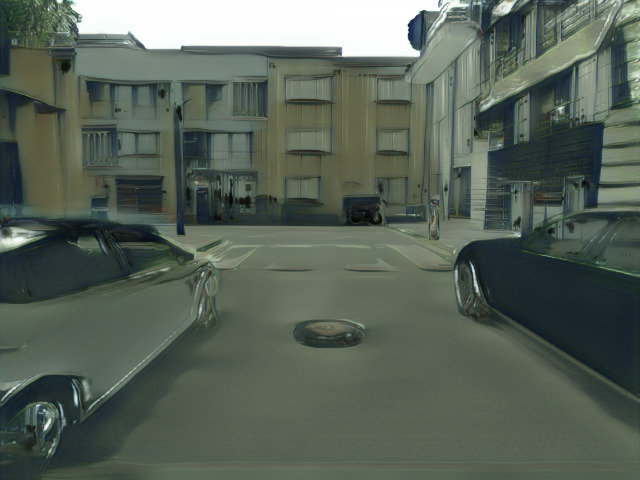}
  \includegraphics[width=\ganwidthapp]{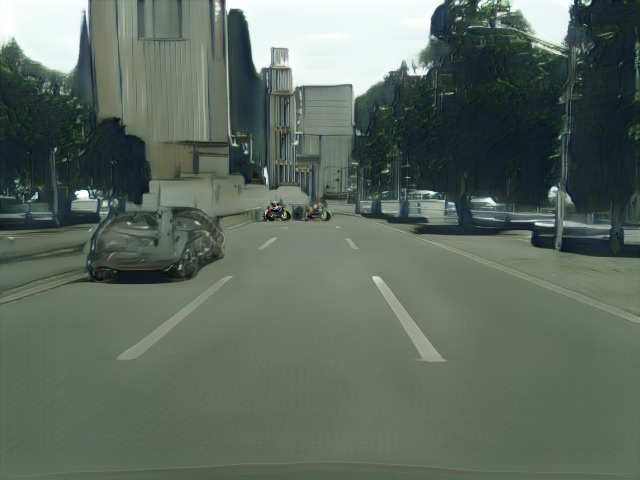} &
  \includegraphics[width=\ganwidthapp]{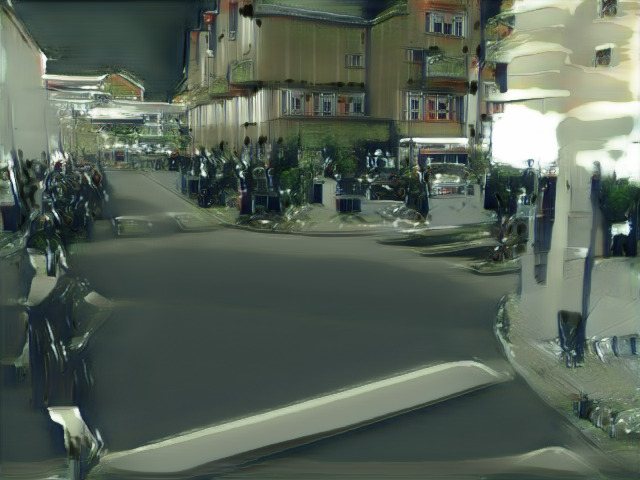} 
  \includegraphics[width=\ganwidthapp]{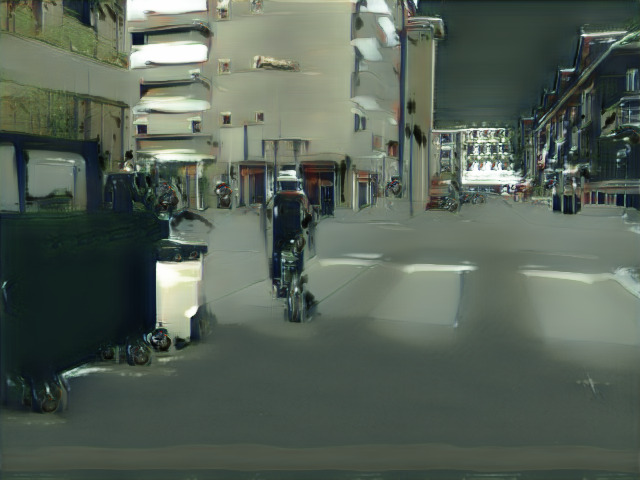} \\
  \rotatebox{90}{GANtruth} \rotatebox{90}{(S+D+I)} &
  \includegraphics[width=\ganwidthapp]{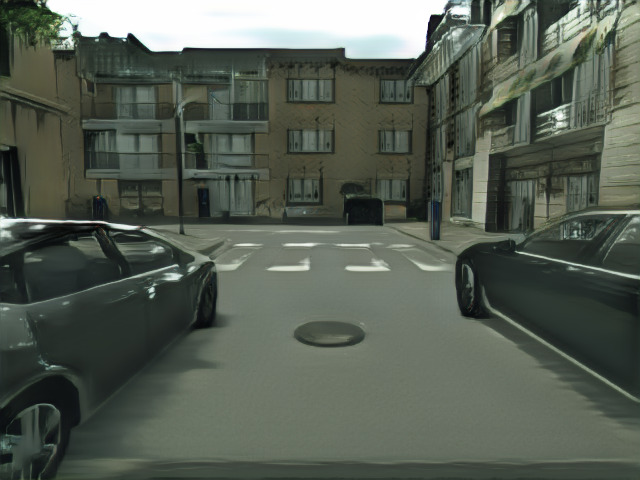}
  \includegraphics[width=\ganwidthapp]{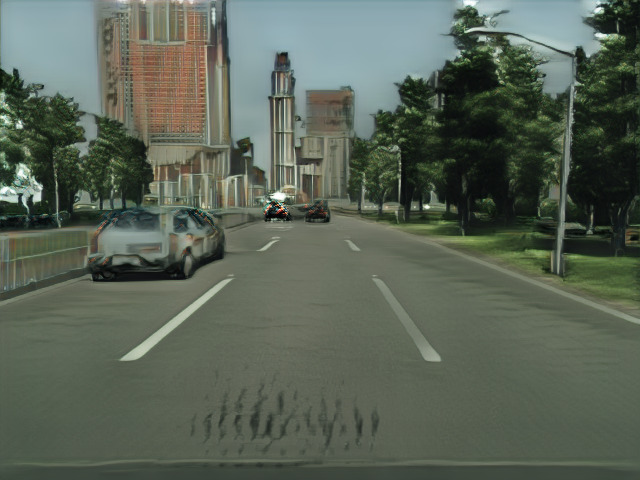} &
  \includegraphics[width=\ganwidthapp]{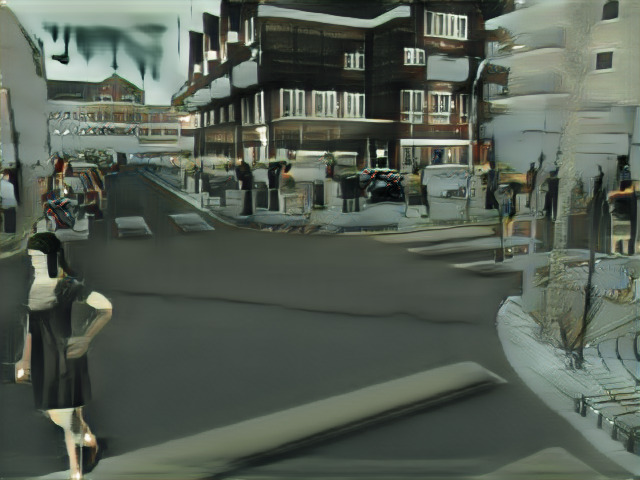} 
  \includegraphics[width=\ganwidthapp]{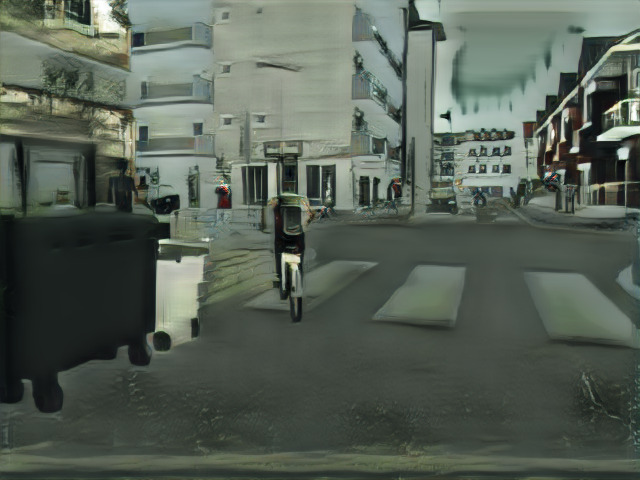} \\
  \rotatebox{90}{UNIT} \rotatebox{90}{(baseline)} &
  \includegraphics[width=\ganwidthapp]{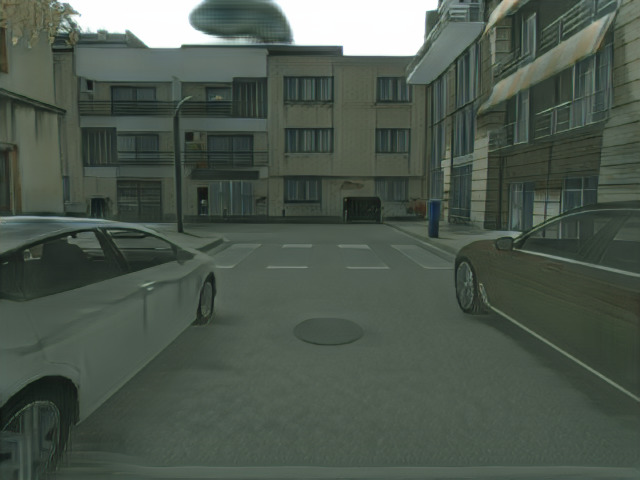}
  \includegraphics[width=\ganwidthapp]{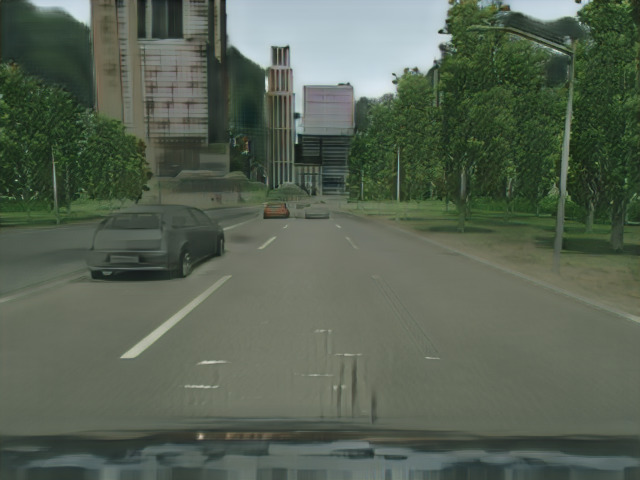} &
  \includegraphics[width=\ganwidthapp]{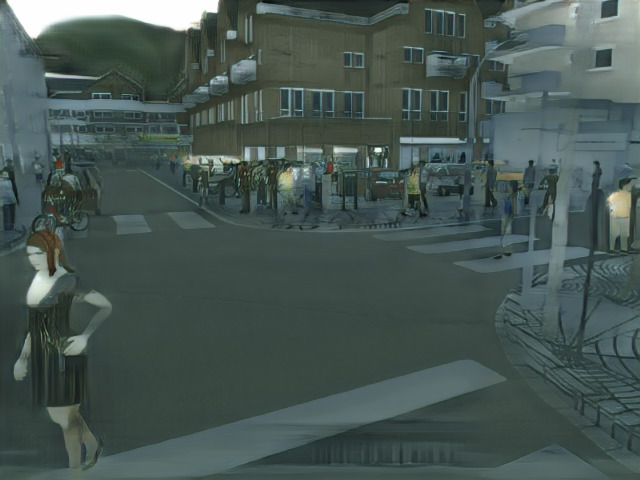} 
  \includegraphics[width=\ganwidthapp]{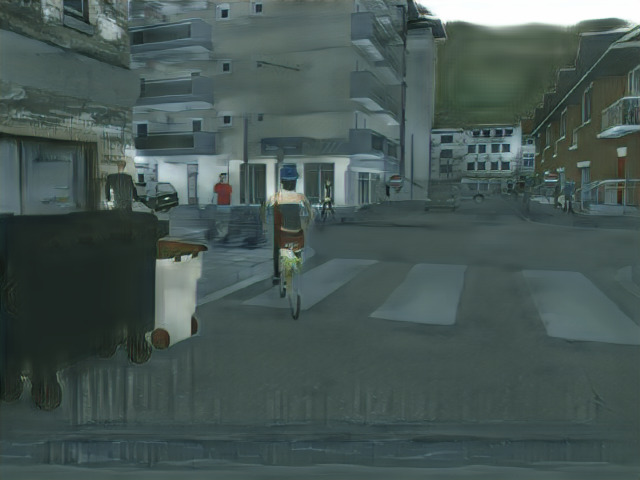} \\
  \rotatebox{90}{UNIT+} \rotatebox{90}{GANtruth} \rotatebox{90}{(S+D)}&
  \includegraphics[width=\ganwidthapp]{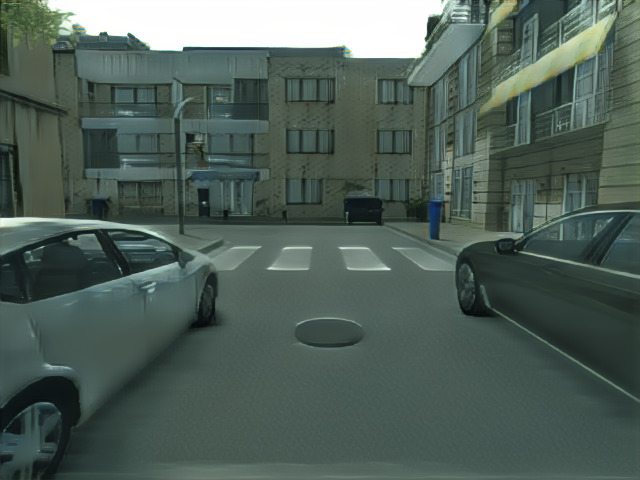}
  \includegraphics[width=\ganwidthapp]{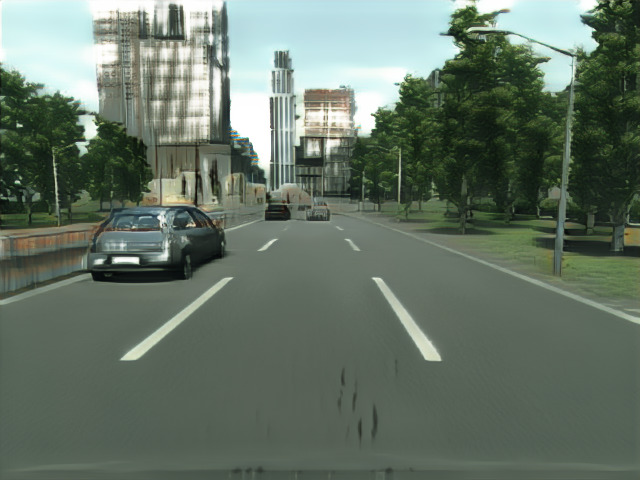} &
  \includegraphics[width=\ganwidthapp]{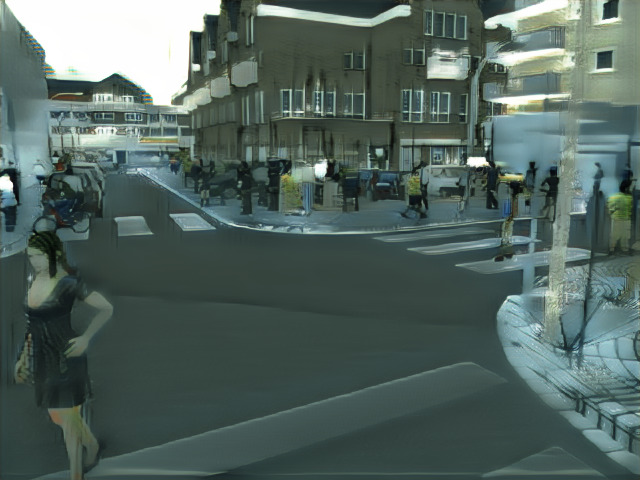} 
  \includegraphics[width=\ganwidthapp]{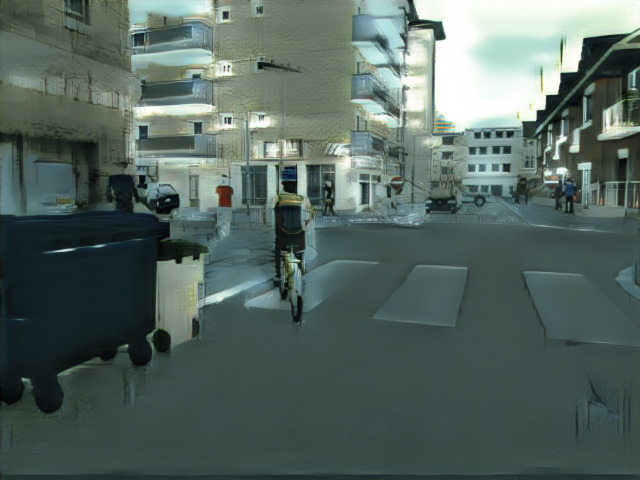} \\
  \rotatebox{90}{UNIT+} \rotatebox{90}{GANtruth} \rotatebox{90}{(S+D+I)}&
  \includegraphics[width=\ganwidthapp]{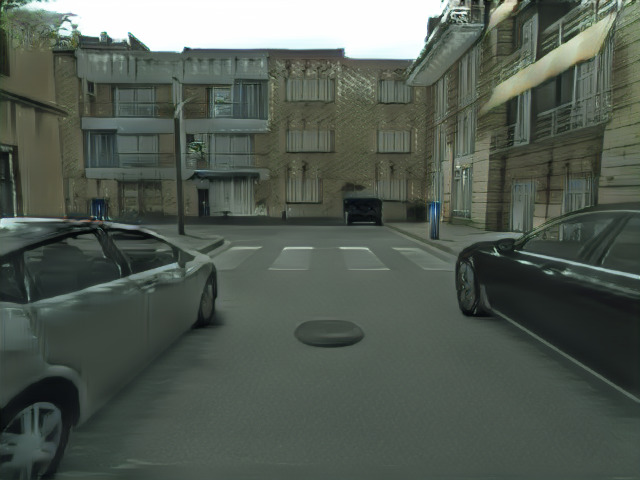}
  \includegraphics[width=\ganwidthapp]{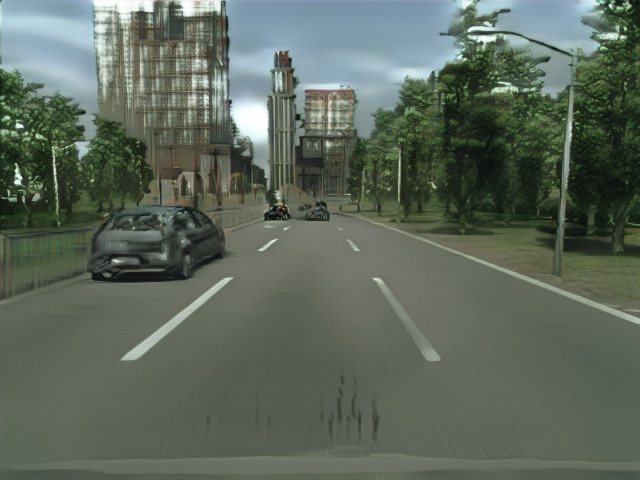} &
  \includegraphics[width=\ganwidthapp]{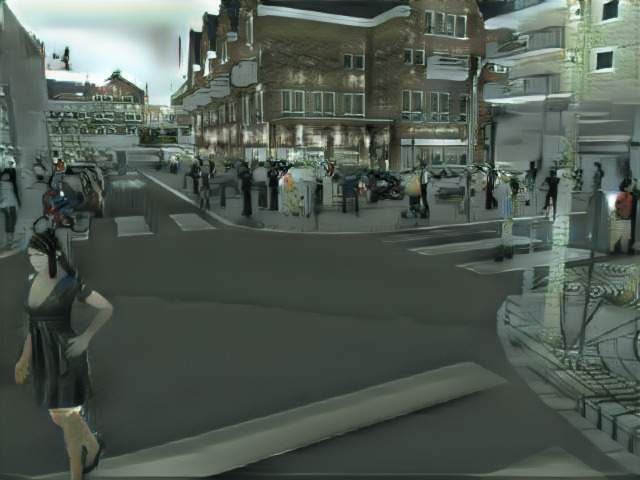} 
  \includegraphics[width=\ganwidthapp]{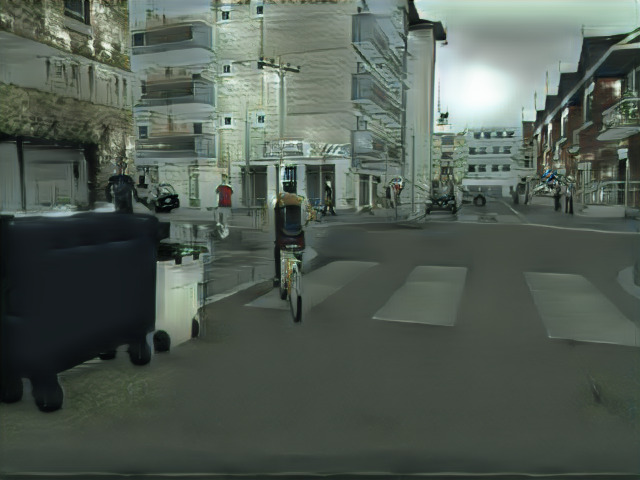} \\
 \end{tabular}
 \caption{Qualitative results. Images randomly sampled. S = Semantic Segmentation, D = Depth, I = Instance Segmentation.}
 \label{fig:appendix_qualitative_results}
\end{figure}

\section{Additional adaptation results}
\label{app:additional_adaptation_results}

\subsection{Adaptation for semantic segementation}

More results from experiments described in Section~\ref{sec:semseg_adaptation} are shown in Table~\ref{tbl:semseg_evaluation_full}.
A possible explanation of why our observations differ from results presented in SimGAN \cite{shrivastava2017learning} is that in driving datasets, the gap between domains is much higher.
This is evident when comparing the accuracies of models trained on source versus target domains. The difference between them was only marginal on the datasets used by \citet{shrivastava2017learning}, while at the same time the amount of synthetic data was greatly exceeding the real data.

\begin{table}[tb!]
 \caption{Adaptation for semantic segmentation. S = Semantic Segmentation, D = Depth, I = Instance Segmentation.}
 \label{tbl:semseg_evaluation_full}
 \centering
 \begin{tabular}{lc}
  \toprule
  Dataset                                   & mIOU (\%) \\
  \midrule
  Source domain (SYNTHIA-Seq)               & 24.9   \\
  \midrule
  Simple GAN (baseline)                     & 20.3   \\
  \midrule
  GANtruth (S+D)               & \textbf{26.6}   \\
  GANtruth (D+I)              & 21.5   \\
  GANtruth (S+D) + sem.consistency \cite{Hoffman_cycada2017}  & 24.4     \\
  GANtruth (D+I) + sem.consistency \cite{Hoffman_cycada2017}  & 25.2      \\
  \midrule
  UNIT (baseline)                           & 23.0   \\
  UNIT + GANtruth (D+I)       & 22.9      \\
  \midrule
  Target domain (Cityscapes)                  & \textbf{57.0}   \\
  Target domain + GANtruth (S+D) + sem.consistency  & 40.3 \\
  Target domain + GANtruth (D+I) + sem.consistency & 40.6 \\
  \bottomrule
 \end{tabular}
\end{table}

\subsection{Adaptation for monocular depth estimation}

Translated datasets were also used for training unsupervised monocular depth estimation network (the same model as the one used for depth preservation) introduced by \citet{monodepth17}.
We find that even though our method improves the results over the baseline model, the results are worse than training on the source domain dataset.
We believe that this is the case because the task network relies on consistency between left and right images by using a reconstruction loss on the reprojected images.
In our case, the left and right images are translated independently, therefore any inconsistency between them is likely to affect the training.
In Table~\ref{tbl:depth_evaluation} we show results from our observations. The models are evaluated on Kitti dataset as in \citet{monodepth17}. Because of the different focal length and camera baseline in the dataset used in evaluation, we use a metric that we refer to as "scale aligned" absolute relative error. The estimated depth maps are aligned with the ground-truth by multiplying them by a single constant before calculating the absolute relative error. Such modified metric is more informative, because it is more independent of the scale of the disparities that depend on the camera parameters.

\begin{table}[tb!]
 \caption{Adaptation for monocular depth estimation. Evaluation on Kitti dataset. S = Semantic Segmentation, D = Depth.}
 \label{tbl:depth_evaluation}
 \centering
 \begin{tabular}{lc}
  \toprule
  Dataset                                   & "scale aligned"  \\
                                            & abs. rel. error  \\
  \midrule
  Source domain (SYNTHIA-Seq)               & \textbf{0.2801}   \\
  \midrule
  GANtruth (S)                       & 0.3401   \\
  GANtruth (D)                          & 0.2849   \\
  GANtruth (S+D)               & 0.3051   \\
  \midrule
  UNIT                                      & 0.3417   \\
  UNIT + GANtruth (S)                & 0.3170   \\
  UNIT + GANtruth (D)                   & 0.2862   \\
  UNIT + GANtruth (S+D)        & 0.2804   \\
  \bottomrule
 \end{tabular}
\end{table}

\section{Labels mapping}
\label{app:label_mapping}
Even though all datasets we use correspond to driving scenes, semantic classes that they define are not the same.
To address this problem we map the source labels as described in Table~\ref{tbl:labels_mapping_semseg}.
Some classes in SYNTHIA do not have correspondances in Cityscapes and/or Microsoft COCO (denoted as NULL), therefore the ground-truth preservation loss (\ref{eq:label_preservation_loss}) ignores such regions or we mask the gradients from those parts during back-propagation in the case of instance segmentation.

\begin{table}[tb!]
 \caption{Mapping of semantic classes from SYNTHIA to Cityscapes}
 \label{tbl:labels_mapping_semseg}
 \centering
 \begin{tabular}{cccccc}
  \toprule
  \multicolumn{2}{c}{(from) SYNTHIA} & \multicolumn{2}{c}{(to) Cityscapes} & \multicolumn{2}{c}{(to) Microsoft-COCO} \\
  \cmidrule(r){1-2} \cmidrule(r){3-4} \cmidrule(r){5-6} 
  Class ID & Class & Class ID & Class & Class ID & Class  \\
  \midrule
    0 & void & - & NULL & - & NULL \\
    1 & sky & 10 & sky & - & NULL \\
    3 & road & 0 & road & - & NULL \\
    4 & sidewalk & 1 & sidewalk & - & NULL \\
    5 & fence & 4 & fence & - & NULL \\
    6 & vegetation & 8 & vegetation & - & NULL \\
    7 & pole & 5 & pole & - & NULL \\
    8 & car & 13 & car & 3 & car \\
    9 & traffic sign & 7 & traffic sign & 13 & traffic sign \\
    10 & pedestrian & 11 & person & 1 & person \\
    11 & bicycle & 18 & bicycle & 2 & bicycle \\
    12 & lane-marking & 0 & road & - & NULL \\
    13 & reserved & - & NULL & - & NULL \\
    14 & reserved & - & NULL & - & NULL \\
    15 & traffic light & 6 & traffic light & 10 & traffic light \\
  \bottomrule
 \end{tabular}
\end{table}

\newpage
\section{Model architecture}
\label{app:appendix_model_architecture}
\begin{figure}
  \centering
  \subfloat[Encoders -- $E_S$, $E_T$]{
    \includegraphics[width=0.40\textwidth]{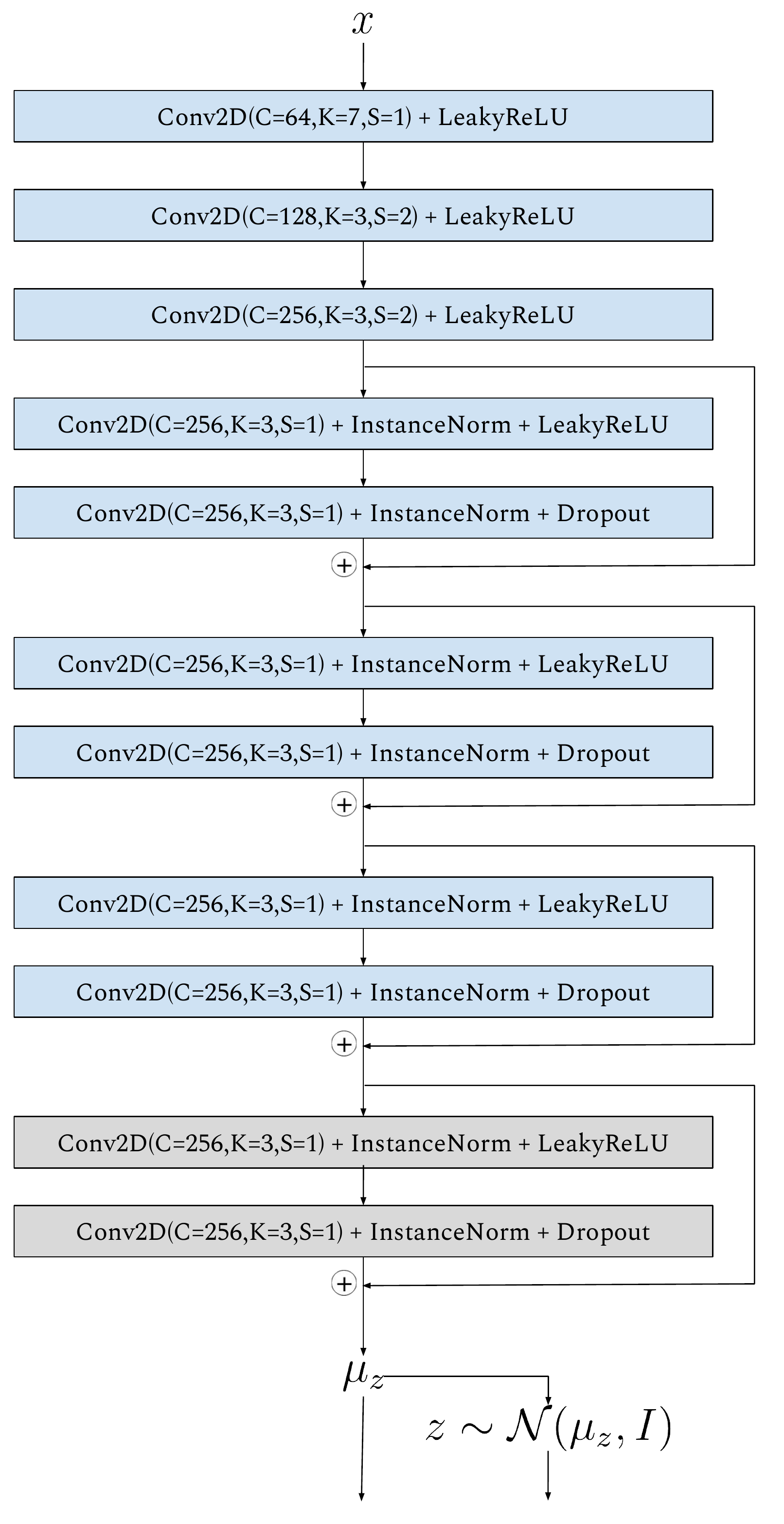}
  }
  \subfloat[Decoders (generators) -- $G_S$, $G_T$]{
    \includegraphics[width=0.40\textwidth]{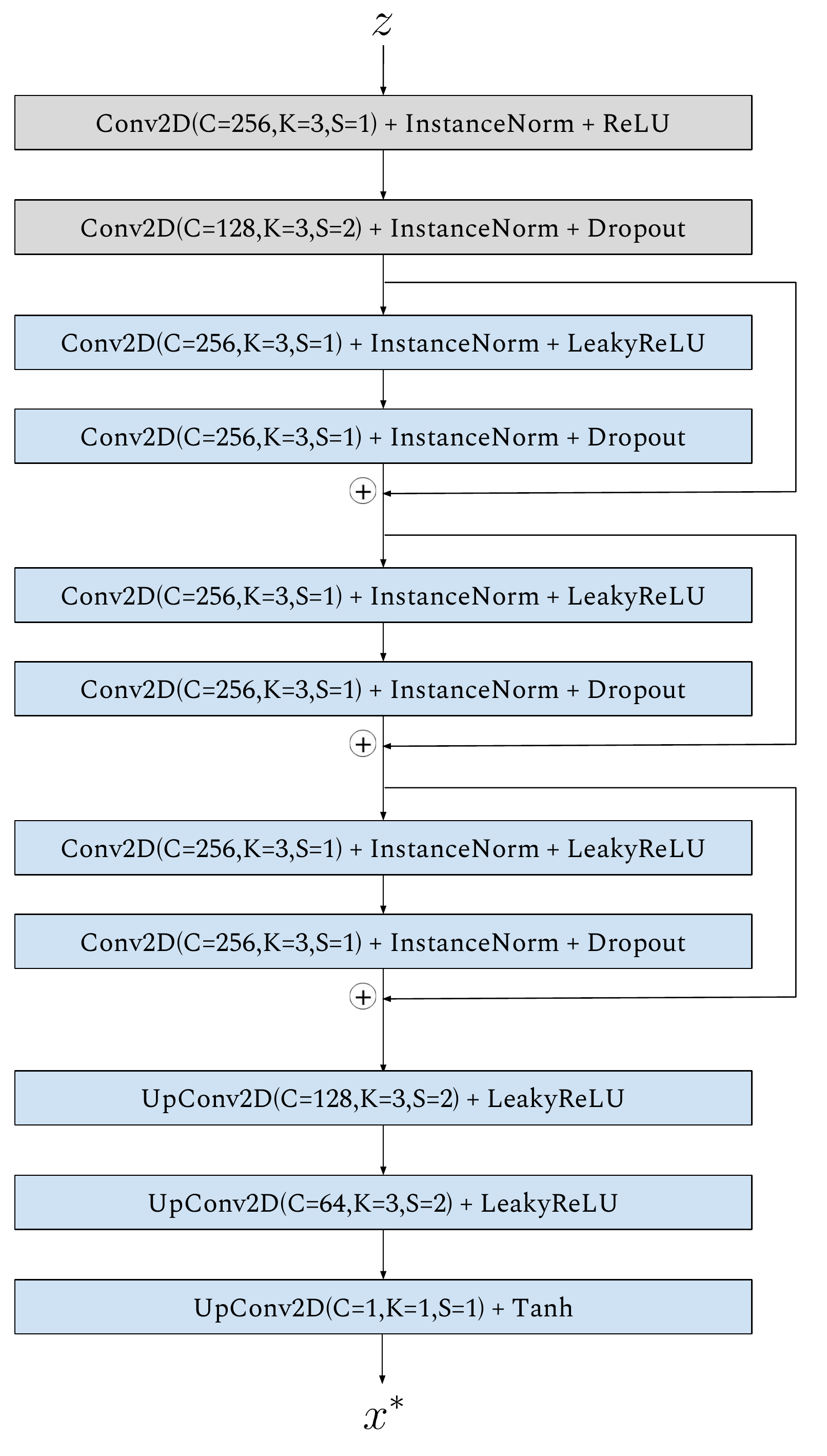}
  }
  \caption{Diagram of encoders and decoders of the models, modules marked as gray have the weights shared between domains. $C$, $K$ and $S$ correspond to the number of channels, kernel size and stride respectively}
  \label{fig:model_architecture}
\end{figure}

\begin{figure}
    \centering
    \includegraphics[width=0.28\textwidth]{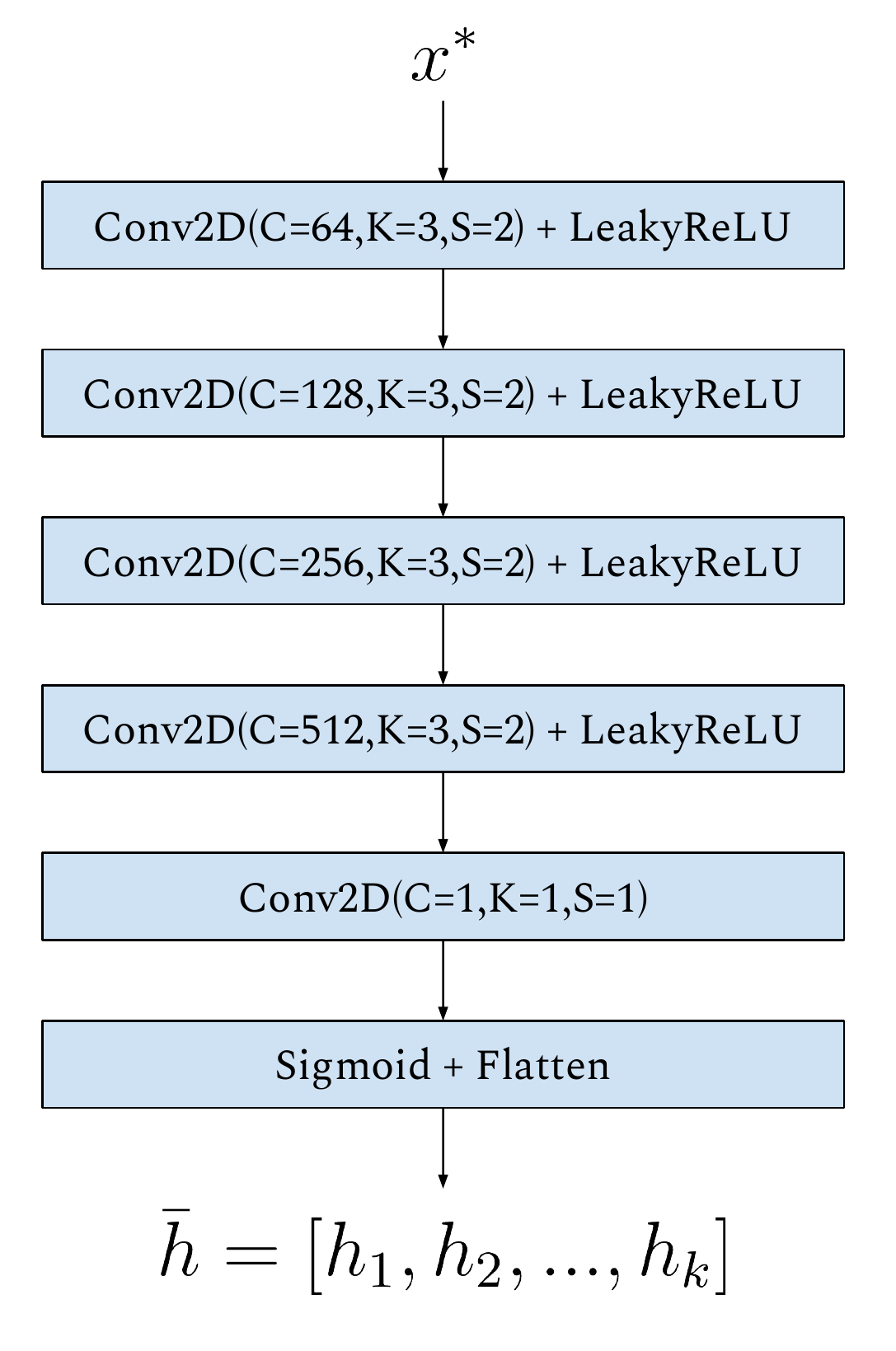}
    \caption{Detailed diagram of each discriminator -- the module does not have any fully connected layers. The output $\bar{h}$ represents a probability on the input image $x^*$ being true or fake, the number of output elements $h_i$ depends on the input resolution. Each $h_i$ has a receptive field of a different patch of the image. We use a multiscale discriminators -- each of them works on the image of different resolution, but the general architecture of the discriminators is the same}
    \label{fig:discriminator_architecture}
\end{figure}

The model architecture used in our experiments corresponds to the one proposed by \citet{liu2017unsupervised} to make the comparisons fair.
The details of the encoders and decoders are present in Figure~\ref{fig:model_architecture}. All of our discriminators are multi-scale. Each of them working on three different resolutions and have the architecture as described in Figure~\ref{fig:discriminator_architecture}.

\end{document}